\documentclass{article} %
\usepackage{style}

\usepackage{amsmath,amsfonts,bm}

\def\eqref#1{equation~\ref{#1}}

\def\1{\bm{1}}

\DeclareMathAlphabet{\mathsfit}{\encodingdefault}{\sfdefault}{m}{sl}
\SetMathAlphabet{\mathsfit}{bold}{\encodingdefault}{\sfdefault}{bx}{n}

\usepackage{fontawesome5}
\usepackage{simpleicons}
\usepackage{url}            %
\usepackage{booktabs}       %
\usepackage{amsfonts}       %
\usepackage{nicefrac}       %
\usepackage{microtype}      %
\usepackage{xcolor}         %
\usepackage{caption}
\usepackage{graphicx}
\definecolor{mydarkblue}{rgb}{0.68, 0.85, 1.0}

\usepackage[utf8]{inputenc} %
\usepackage[T1]{fontenc}    %

\usepackage{url}            %
\usepackage{booktabs}       %
\usepackage{amsfonts}       %
\usepackage{nicefrac}       %
\usepackage{microtype}      %
\usepackage{soul}

\usepackage{sidecap}
\usepackage{float}
\usepackage{graphicx}
\usepackage{subcaption}

\usepackage{mathrsfs}

\usepackage{amsmath}
\usepackage{amssymb}
\usepackage{amsthm}
\usepackage{mathtools}
\usepackage[mathscr]{eucal}%
\usepackage{bm}%
\usepackage{bbm}
\usepackage{relsize}
\usepackage{graphics}
\usepackage{wrapfig}
\usepackage{multirow}
\usepackage{enumitem}
\usepackage{tablefootnote}
\usepackage{makecell}

\usepackage{array}
\usepackage{color, colortbl}
\definecolor{Gray}{gray}{0.9}
\definecolor{darkpastelgreen}{rgb}{0.01, 0.75, 0.24}
\definecolor{cadetgrey}{rgb}{0.57, 0.64, 0.69}
\definecolor{camel}{rgb}{0.76, 0.6, 0.42}
\definecolor{lightskyblue}{rgb}{0.53, 0.81, 0.98}
\definecolor{lightsblue}{rgb}{0.53, 0.81, 0.98}
\definecolor{lightblue}{rgb}{0.68, 0.85, 0.9}
\definecolor{softblue}{rgb}{0.85, 0.91, 0.98}
\definecolor{mygray}{gray}{0.6}

\definecolor{SparseBlue}{HTML}{EEF3FF}
\definecolor{DensePurple}{HTML}{F1E6FF}
\definecolor{FramePurple}{HTML}{C2185B}

\usepackage{pgf}
\usepackage{xinttools}
\usepackage{tikz}
\usetikzlibrary{arrows.meta}
\usepackage{textcomp}
\usetikzlibrary{calc,shapes,arrows,positioning,automata,trees}
\usepackage{placeins} 
\usepackage{tabularx}
\usepackage{graphicx}
\usepackage{subcaption} 
\usepackage{listings}

\usepackage[font=small,labelfont=bf,tableposition=top]{caption}

\DeclareCaptionLabelFormat{andtable}{#1~#2  \&  \tablename~\thetable}

\usepackage{blindtext}

\usepackage{bigdelim}
\usepackage{colortbl} %
\definecolor{mColor1}{rgb}{0.95,0.95,0.95}

\newcommand{\score}[1]{\makebox[2.4em][l]{#1}}
\newcommand{\gain}[2]{\makebox[2.4em][l]{#1}{\scriptsize #2}}
\newcommand{\bgain}[2]{\makebox[2.4em][l]{\textbf{#1}}{\scriptsize #2}}

\usepackage{microtype}
\usepackage{graphicx}
\usepackage{booktabs} %
\usepackage{soul}

\definecolor{mydarkblue}{rgb}{0.68, 0.85, 1.0}
\definecolor{mydarkblue2}{rgb}{0,0.08,0.45}
\definecolor{mydarkblue3}{RGB}{151,204,255}
\definecolor{cvprblue}{rgb}{0.21,0.49,0.74}

\usepackage[colorlinks=true,
    linkcolor=mydarkblue2,
    citecolor=cvprblue,
    filecolor=cvprblue,
    urlcolor=mydarkblue2]{hyperref}

\hypersetup{pdftitle={Less from More: Reinforcing Sparse Video Reasoning from Dense References},
  pdfauthor={Wenfang Sun, Yingjun Du, Cees G. M. Snoek}}
\begin{document}
\begin{wtctitleblock}
\wtctitle{Less from More}{Reinforcing Sparse Video Reasoning from Dense References}
\begin{center}
{\sffamily\bfseries
Wenfang Sun\textsuperscript{1}\quad
Yingjun Du\textsuperscript{1,$\dagger$}\quad
Cees G. M. Snoek\textsuperscript{1}\par}
\smallskip
{\sffamily\small\textsuperscript{1}University of Amsterdam\par}
\end{center}

\begin{abstract}
Video-language models commonly assume that more temporal observations lead to more reliable reasoning.
We question this assumption and argue that the key challenge is not merely processing more video frames efficiently, but learning to reason reliably under limited temporal evidence.
We propose \textsc{SAVER}, a dense-to-sparse post-training framework that uses dense video views as training-time references for sparse-frame inference.
During reinforcement post-training, paired dense and sparse views are optimized with grounding rewards and a reliability-gated reference reward, encouraging sparse view predictions to preserve task-relevant temporal evidence.
Notably, \textsc{SAVER} is trained only on 1,250 randomly sampled temporal grounding examples, without using any video question answering annotations.
Across three temporal grounding benchmarks and six video question-answering benchmarks, \textsc{SAVER} consistently improves performance across frame budgets.
In particular, \textsc{SAVER} can match or surpass dense-frame Qwen3.5 baselines while using substantially fewer frames.
These results show that temporal grounding can serve as an effective evidence-localization proxy for learning sparse video reasoning that transfers to broader video understanding tasks.
\end{abstract}

\wtctitlefooter{%
  \textsuperscript{$\dagger$}Corresponding author.\\[4pt]
  {\fontsize{9}{11}\selectfont
    \href{https://wenfangsun.cn/SAVER/}{%
      \textcolor{accent}{\faGlobe}~\textbf{Project Page}}
    \qquad
    \href{https://github.com/lmsdss/SAVER}{%
      \textcolor{accent}{\simpleicon{github}}~\textbf{GitHub}}
    \qquad
    \href{https://huggingface.co/collections/lmsdss/saver-less-from-more}{%
      \textcolor[HTML]{FFD21E}{\simpleicon{huggingface}}~\textbf{Models}}
  }%
}

\end{wtctitleblock}
\begingroup
    \centering
    \includegraphics[width=1\linewidth]{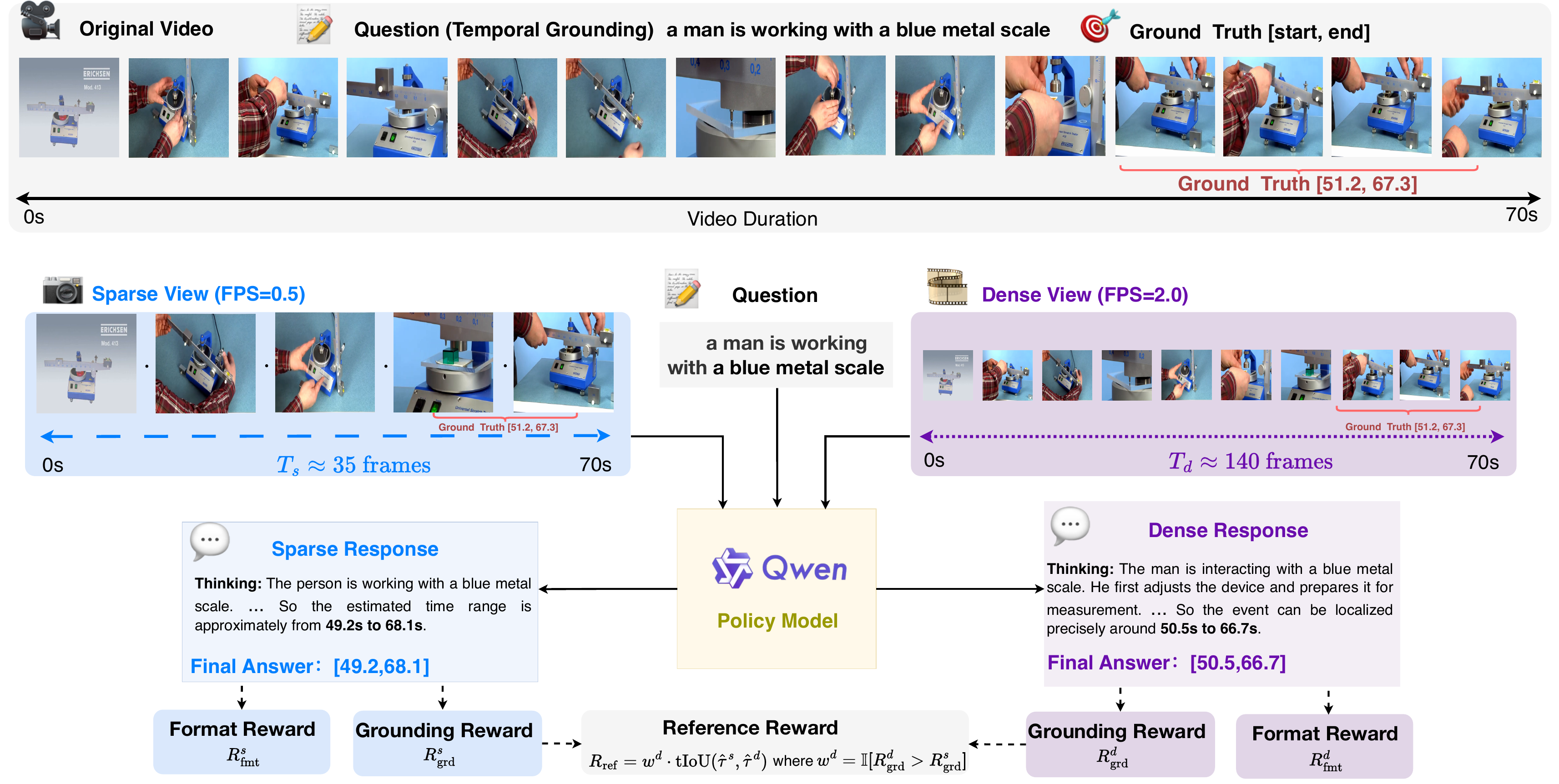}
\captionof{figure}{
\textbf{Overview of \textsc{SAVER}.}
\textsc{SAVER} performs dense-to-sparse post-training with paired views from the same video.
Dense views provide training-time references, while sparse views expose the policy to reduced temporal observations.
Format and grounding rewards supervise both views, and a reliability-gated reference reward aligns sparse predictions with reliable dense predictions.
At inference time, the model uses sparse frames.
}
\label{fig:saver_overview}

\par\endgroup
\clearpage

\section{Introduction}

A common recipe for improving video reasoning is to show multimodal large language models (MLLMs) more frames.
Recent MLLM families, including Qwen-VL~\citep{bai2023qwenvl,qwen2vl,qwen2_5_vl,bai2025qwen3,qwen35blog}, InternVL~\citep{chen2024internvl,chen2024expanding}, and LLaVA-style models~\citep{liu2024visual,li2024llavaonevision, lin2023video}, have substantially strengthened visual understanding, while video post-training further enhances reasoning on question answering, perception, and temporal tasks~\citep{feng2026video,liu2026videoauto,feng2026onethinker}.
Yet most video reasoning pipelines still rely on a dense-input assumption: reliable predictions require observing many frames.
This assumption is costly, as dense temporal inputs can involve tens or hundreds of frames, leading to large visual-token budgets, high memory usage, and increased latency.
Such costs are especially problematic for streaming video, mobile assistants, robotics, surveillance, and long-form video search, where models must operate under limited compute or latency budgets.
Efficient video reasoning therefore requires reducing the visual evidence processed in the first place, not merely decoding responses faster.

This raises a simple but underexplored question:
\emph{do video-language models truly need dense frame sampling at inference time, or can they be post-trained to reason reliably from sparse temporal observations?}
The answer is non-trivial. Simply reducing the frame rate lowers visual cost, but it also removes temporal evidence that may be necessary for answering questions or localizing events~\citep{li2025improving}. At the same time, dense sampling often introduces substantial redundancy, and its computational cost grows quickly with the number of observed frames. The central challenge we address in this paper is therefore not merely to process dense videos more efficiently, but to make sparse-frame inference itself reliable.

\begingroup
\setlength{\intextsep}{4pt}
\begin{figure}[H]
    \centering
    \includegraphics[width=1.\linewidth]{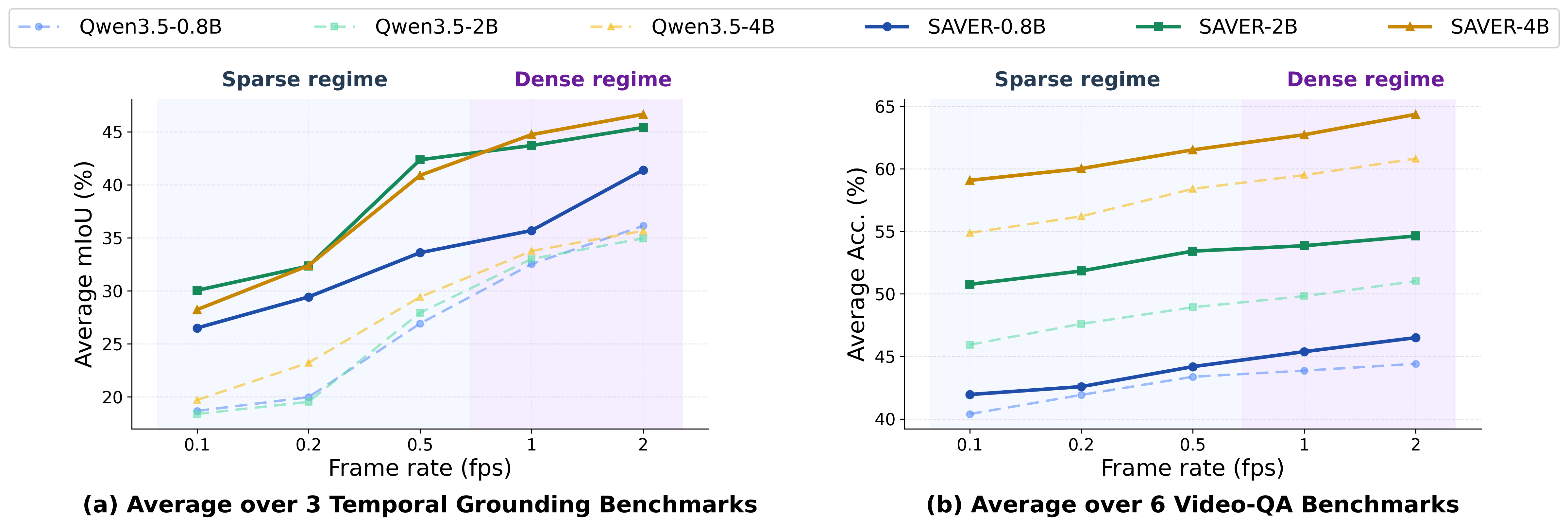}
\caption{
\textbf{Performance--efficiency trade-off across frame rates.}
Lower frame rates reduce the number of sampled frames but enlarge the sparse--dense performance gap for Qwen3.5. 
\textsc{SAVER} mitigates this degradation; for example, for video-QA, \textsc{SAVER}-2B at 0.1 fps nearly matches dense Qwen3.5-2B at 2.0 fps.
}
    \label{fig:intro_fps_tradeoff}
    
\end{figure}
\endgroup

Figure~\ref{fig:intro_fps_tradeoff} illustrates the accuracy--frame-budget trade-off.
Lowering the frame rate reduces the number of sampled frames substantially, but it also degrades standard Qwen3.5 models, especially on temporal grounding where accurate localization depends on sufficient temporal evidence.
\textsc{SAVER} changes this trade-off: at the same frame rate, it consistently improves over Qwen3.5, and at lower frame rates it can approach the accuracy of denser Qwen3.5 inference while using far fewer frames.
For example, on video-QA benchmarks, \textsc{SAVER}-2B at 0.1 fps nearly matches Qwen3.5-2B at 2.0 fps.
Importantly, this improvement transfers from temporal grounding, the task used for post-training, to six video-QA benchmarks, suggesting that learning to preserve temporal evidence under sparse observations benefits broader video reasoning.

Existing efficiency approaches only partially address this problem.
Some methods accelerate reasoning on the language side, for example by reducing redundant decoding or reasoning steps~\citep{liu2026videoauto}, but they do not directly reduce the cost of visual input processing.
Other methods reduce visual computation through token compression~\citep{wu2025marc}, adaptive keyframe sampling~\citep{tang2025adaptive}, or retrieval-based frame selection~\citep{zou2026videobrain}.
These approaches are effective, but they typically assume a richer visual input and then reduce its cost by compressing, selecting, or retrieving visual evidence.
Our work addresses a different part of the efficiency problem: given a prescribed low-frame-rate visual budget, can the video-language model itself be made reliable under sparse temporal evidence?
In this work, we study \emph{sparse video reasoning}: the model receives only sparsely sampled frames at inference time and must reason reliably under limited temporal evidence.
Sparse observations may miss key events, provide coarse temporal boundaries, or emphasize misleading static cues, making models trained mainly under dense observations fragile when the frame budget is reduced.
Rather than designing another frame selector or token compressor, we take a training-centric view: dense frames should serve as training-time references for learning sparse-frame reasoning, but should not be required at inference time. Motivated by this view, we propose \textsc{SAVER}, a dense-guided sparse post-training framework for efficient video understanding.
\textsc{SAVER} constructs paired dense and sparse views from the same video and uses temporal grounding as an evidence-localization proxy task.
Dense views provide richer temporal reference signals, while sparse view predictions are optimized with grounding rewards and a reference reward that aligns them with reliable dense view predictions.

Notably, \textsc{SAVER} is trained only on 1,250 randomly sampled temporal grounding examples from Time-R1~\citep{wang2025time}, without using any video-QA annotations.
Despite this small grounding-only training set, \textsc{SAVER} consistently improves across three temporal grounding benchmarks and six video question-answering benchmarks under different frame budgets, with particularly strong gains in the sparse regime.
These results suggest that efficient video reasoning should not only focus on compressing dense-frame inference, but also on training models to remain reliable when dense temporal evidence is unavailable.

\section{Related Work}

\noindent\textbf{Visual reasoning models.}
Recent multimodal large language models (MLLMs) have substantially advanced image and video understanding. 
Representative model families include Qwen-VL~\citep{bai2023qwenvl,qwen2vl,qwen2_5_vl,bai2025qwen3,qwen35blog}, InternVL~\citep{chen2024internvl,chen2024expanding}, LLaVA-style models~\citep{liu2024visual,li2024llavaonevision}, and video-oriented MLLMs such as Video-LLaVA~\citep{lin2023video},
VideoChat~\citep{li2025videochat},
and TimeChat~\citep{ren2024timechat}.
Building on these models, reinforcement post-training has become a popular strategy for improving visual reasoning.
VisionReasoner~\citep{liu2026visionreasoner} and RegionReasoner~\citep{sun2026regionreasoner} focus on image-level and region-grounded reasoning, while Video-R1~\citep{feng2026video}, VideoChat-R1~\citep{li2025videochatr1}, VideoAuto-R1~\citep{liu2026videoauto}, and OneThinker~\citep{feng2026onethinker} extend reasoning-oriented post-training to video.
Other works further investigate temporal grounding~\citep{wang2025time,li2025reinforcement}, video reasoning segmentation~\citep{gong2025reinforcing}, frame-referenced reasoning traces~\citep{ghazanfari2025chain}, and scalable video reasoning training~\citep{chen2025scaling,li2025veripo,fu2025love}.
These methods improve the model's reasoning behavior or supervision signal, but generally assume that sufficient visual evidence is available at inference time. In contrast, our work studies whether video reasoning can remain reliable when only sparse temporal observations are provided at test time. 

\noindent\textbf{Efficient video understanding.}
The high computational cost of video understanding has motivated methods that reduce either language-side or visual-side computation. 
Language-side methods reduce redundant reasoning or decoding steps; for example, VideoAuto-R1~\citep{liu2026videoauto} avoids unnecessary repeated reasoning, while Sparse-to-Dense ~\citep{zhang2025sparse} accelerates decoding through sparse-attention speculation and dense-attention verification.
Visual-side methods reduce the cost of processing long or dense videos by compressing visual tokens or selecting informative frames. 
MARC~\citep{wu2025marc} studies memory-augmented token compression for efficient video understanding, STORM~\citep{jiang2025storm} reduces visual tokens for long-video MLLMs, and frame selection or retrieval methods such as M-LLM-based frame selection~\citep{hu2025mllm}, Adaptive Keyframe Sampling~\citep{tang2025adaptive}, Generative Frame Sampler~\citep{yao2025generative}, VideoBrain~\citep{zou2026videobrain}, and ReViSe~\citep{Xu_2026_CVPR} select or retrieve useful frames from richer video contexts. Q-Frame~\citep{zhang2025q} further combines query-aware frame selection with multi-resolution adaptation.  FastVID~\citep{shen2026fastvid} dynamically prunes redundant visual information, while WFS-SB~\citep{Chen_2026_CVPR} selects informative frames by detecting semantic boundaries.
These works show that not all frames or visual tokens contribute equally to video understanding. 
Our setting differs in that we evaluate the model under a \emph{fixed low-frame-rate input budget}: the model receives only uniformly sampled sparse frames at inference time, rather than first accessing a richer video representation and then selecting or compressing it.
Accordingly, \textsc{SAVER} does not propose a new frame selector or token compressor; instead, it uses dense views only during training as scaffolds and aligns sparse view predictions with dense view signals through reinforcement post-training. 
This targets a different question: how can a video model remain reliable when dense temporal evidence is unavailable or too costly to process at inference time?

\section{Method}
\label{sec:method}

\textsc{SAVER} targets efficient video inference under a low-frame-rate protocol.
At test time, the model receives frames at a prescribed fps and predicts from this sparse input.
It does not perform dense video pre-scanning, adaptive frame selection, or test-time optimization.
Thus, the goal is not to introduce a sampling strategy, but to make the model reliable when only sparse frames are processed.

\subsection{Overview}

For each training example, \textsc{SAVER} follows three simple steps.
First, we sample two views from the same video: a dense view $V^d$ with more frames and a sparse view $V^s$ with fewer frames.
Second, both views are fed into the same MLLM policy $\pi_\theta$, producing a dense view response $y^d$ and a sparse view response $y^s$.
Third, we optimize both responses with temporal grounding rewards and add a dense-to-sparse reference reward, so that the sparse prediction learns from the dense prediction only when the dense prediction is more reliable.
After training, only the sparse view is used for inference.

Given a video $V$ and a temporal grounding instruction $q$, we construct:
\begin{equation}
    V^d = \mathcal{S}_d(V), 
    \qquad
    V^s = \mathcal{S}_s(V),
\end{equation}
where $\mathcal{S}_d$ and $\mathcal{S}_s$ denote dense and sparse frame sampling functions, respectively.
Both views are fed into the same MLLM policy:
\begin{equation}
    y^{d} \sim \pi_\theta(\cdot \mid V^{d}, q), 
    \qquad
    y^{s} \sim \pi_\theta(\cdot \mid V^{s}, q).
\end{equation}
During post-training, each response is expected to predict a temporal interval.
Although \textsc{SAVER} is trained only with temporal grounding supervision, we later evaluate the resulting sparse view policy on both temporal grounding and video question answering to test cross-task transfer.

\subsection{Dense-to-Sparse Post-Training}
\label{sec:grounding_training}
\textsc{SAVER} performs reinforcement post-training with paired dense and sparse video views, where dense views provide training-time guidance for improving sparse-frame inference.
We use temporal grounding as the training task because it provides explicit supervision over when an event occurs, making it a natural signal for learning task-relevant temporal evidence.
In our implementation, we randomly sample 1,250 training examples from Time-R1~\citep{wang2025time}.
No video-QA annotations are used during post-training.
Therefore, improvements on video-QA benchmarks come from cross-task transfer rather than task-specific QA post-training.

Given a video-instruction pair $(V,q)$ and a ground-truth temporal segment
$\tau^\star=(t_s^\star,t_e^\star)$, the model predicts temporal segments from both views:
\begin{equation}
    \hat{\tau}^{d}=(\hat{t}^{d}_{s},\hat{t}^{d}_{e})
    \leftarrow y^d,
    \qquad
    \hat{\tau}^{s}=(\hat{t}^{s}_{s},\hat{t}^{s}_{e})
    \leftarrow y^s.
\end{equation}
The dense and sparse responses are optimized with temporal grounding rewards.
We then use the dense prediction as a training-time reference for the sparse prediction only when the dense view is more reliable.
This dense-to-sparse alignment encourages the sparse view to preserve temporal evidence that is available in the dense view, while avoiding imitation of unreliable dense predictions.

\subsection{Reward Design}
\label{sec:reward}

\textsc{SAVER} uses three rewards during post-training: a format reward that ensures valid temporal outputs, a grounding reward that evaluates each predicted temporal interval against the ground truth, and a reference reward that transfers reliable dense view signals to the sparse view.

\noindent\textbf{Format reward.}
The format reward $R_{\mathrm{fmt}}(y)$ encourages the model to generate a valid and parseable temporal grounding response.
It checks whether the response contains a start timestamp and an end timestamp in the required format.
This reward stabilizes reinforcement post-training by preventing malformed outputs.

\noindent\textbf{Grounding reward.}
The grounding reward evaluates the temporal segment predicted by each view.
Given a predicted segment $\hat{\tau}=(\hat{t}_{s},\hat{t}_{e})$ and the ground-truth segment $\tau^\star=(t_s^\star,t_e^\star)$, we define:
\begin{equation}
    R_{\mathrm{grd}}(y,\tau^\star)
    =
    \mathrm{tIoU}(\hat{\tau}, \tau^\star)
    -
    \lambda_b
    \left(
    \frac{|\hat{t}_{s}-t_s^\star|}{T}
    +
    \frac{|\hat{t}_{e}-t_e^\star|}{T}
    \right),
\end{equation}
where $\mathrm{tIoU}$ denotes temporal intersection-over-union, $T$ is the video duration, and $\lambda_b$ controls the boundary penalty.
The tIoU term encourages the predicted segment to overlap with the target event, while the boundary term provides denser feedback for accurately localizing the start and end timestamps.

\noindent\textbf{Reference reward.}
The grounding reward supervises each view independently.
We additionally introduce a reference reward that encourages agreement between the predicted temporal intervals when the dense prediction receives a higher grounding reward.
We therefore introduce a reference reward.
Let $\hat{\tau}^{d}$ and $\hat{\tau}^{s}$ be the temporal intervals predicted from the dense and sparse views, respectively.
We define:
\begin{equation}
    R_{\mathrm{ref}}
    =
    w^d \cdot \mathrm{tIoU}(\hat{\tau}^{s},\hat{\tau}^{d}),
\end{equation}
where $w^d$ is a relative reliability gate:
\begin{equation}
    w^d =
    \mathbb{I}
    \left[
    R_{\mathrm{grd}}(y^d,\tau^\star)
    >
    R_{\mathrm{grd}}(y^s,\tau^\star)
    \right].
\end{equation}
The gate activates the reference reward only when the sampled dense prediction receives a higher grounding reward than its paired sparse prediction.
Otherwise, the reference reward is zero.
This is a relative, sample-level reward-shaping criterion: it neither certifies the correctness of the dense prediction nor guarantees that the combined reward increases monotonically with grounding quality.

\subsection{Training Objective and Inference}
\label{sec:objective}

The dense view reward contains the format reward and the grounding reward:
\begin{equation}
    R^d
    =
    \lambda_{\mathrm{fmt}} R_{\mathrm{fmt}}(y^d)
    +
    \lambda_{\mathrm{grd}} R_{\mathrm{grd}}(y^d,\tau^\star).
\end{equation}
The sparse view reward additionally includes the reference reward:
\begin{equation}
    R^s
    =
    \lambda_{\mathrm{fmt}} R_{\mathrm{fmt}}(y^s)
    +
    \lambda_{\mathrm{grd}} R_{\mathrm{grd}}(y^s,\tau^\star)
    +
    \lambda_{\mathrm{ref}} R_{\mathrm{ref}}.
\end{equation}
We optimize the model with reinforcement post-training over paired dense and sparse views:
\begin{equation}
    \max_{\theta}
    \;
    \mathbb{E}\!\left[
        R^d + \lambda_s R^s
        -
        \beta \mathcal{D}_{\mathrm{KL}}
    \right],
\end{equation}
where $\lambda_s$ controls the relative weight of the sparse view reward and $\beta$ controls KL regularization to the reference policy $\pi_{\mathrm{ref}}$.
The KL term is defined as:
\begin{equation}
    \mathcal{D}_{\mathrm{KL}}
    =
    D_{\mathrm{KL}}\!\left(
    \pi_\theta(\cdot \mid V^d,q)
    \,\|\, 
    \pi_{\mathrm{ref}}(\cdot \mid V^d,q)
    \right)
    +
    D_{\mathrm{KL}}\!\left(
    \pi_\theta(\cdot \mid V^s,q)
    \,\|\, 
    \pi_{\mathrm{ref}}(\cdot \mid V^s,q)
    \right).
\end{equation}
In practice, we follow the reinforcement post-training recipe of Time-R1~\citep{wang2025time} and train on 1,250 randomly sampled temporal grounding examples.
We pair $G=4$ dense and sparse rollouts index-wise and normalize advantages over their combined rewards.
At inference time, only the sparse view is used.
Given a prescribed frame rate, the input video is sampled into a sparse frame sequence 
$V^s=\mathcal{S}_s(V)$, and the model directly predicts $y^s \sim \pi_\theta(\cdot \mid V^s,q)$.

No dense video pre-scanning, adaptive frame selection, or test-time optimization is performed.
This makes the inference protocol compatible with low-fps streaming or resource-constrained settings, where only a sparse frame stream is decoded or processed.

\noindent\textbf{Relation to distillation and risk intuition.}
\textsc{SAVER} is related to self-distillation, but differs in both the input views and the use of the teacher signal.
Conventional self-distillation usually transfers predictions under the same input distribution, whereas \textsc{SAVER} performs reward-guided cross-view distillation from dense temporal views to sparse temporal views.
The dense view provides privileged training-time information, while the sparse view matches the inference-time condition.
Importantly, dense predictions are not treated as unconditional pseudo-labels: the alignment reward is activated only when the dense view grounding reward is higher than the sparse view grounding reward.
This prevents sparse predictions from imitating inferior dense view outputs.

This view also gives a simple risk-decomposition intuition for dense-to-sparse alignment.
Let $p_s(y)=\pi_\theta(y\mid V^s,q)$ and $p_d(y)=\pi_\theta(y\mid V^d,q)$ denote the predictive distributions under sparse and dense views, respectively.
For a bounded task loss $\ell(y,y^\star)\in[0,1]$, the sparse view risk can be upper-bounded by:
\begin{equation}
    \mathbb{E}_{y\sim p_s}[\ell(y,y^\star)]
    \leq
    \mathbb{E}_{y\sim p_d}[\ell(y,y^\star)]
    +
    D_{\mathrm{TV}}(p_s,p_d),
\end{equation}
where $D_{\mathrm{TV}}$ is the total variation distance between the two predictive distributions.
This inequality suggests that sparse view performance depends on two factors: the reliability of dense view predictions and the discrepancy between sparse view and dense view predictions.
\textsc{SAVER} directly targets both factors.
The grounding rewards improve the quality of predictions under each view, while the relative reliability-gated reference reward reduces the sparse--dense discrepancy only when the dense view provides a better temporal reference.
Thus, dense frames are used not to increase test-time computation, but to provide reliable training-time guidance for sparse-frame inference.

\section{Experiments}

\subsection{Experiment Details}

\noindent\textbf{Implementation details.}
We instantiate \textsc{SAVER} on Qwen3.5~\citep{qwen35blog} models with three different scales:
0.8B, 2B, and 4B. 
All models are trained with the same dense--sparse reinforcement post-training framework described in Sec.~\ref{sec:method}. 
Unlike large-scale video reasoning post-training methods, \textsc{SAVER} uses only a small temporal grounding training set. 
Specifically, we randomly sample 1,250 examples from Time-R1~\citep{wang2025time} and use them as the only post-training data. 
No video-QA annotations are used during post-training, allowing us to evaluate whether sparse temporal alignment learned from grounding supervision transfers to broader video understanding tasks. For each training example, we construct a dense view and a sparse view from the same video. 
Both views are fed into the same policy model, and the model is optimized with grounding rewards and dense-to-sparse reference rewards. 
We set the coefficient of the sparse view reward to $2.0$ to emphasize sparse-frame robustness, and set all other reward coefficients to $1.0$. 
For reinforcement learning, the rollout size is set to $G=4$. 
The global batch sizes are set to 64, 64, and 128 for the 0.8B, 2B, and 4B models. 
We train the 0.8B, 2B, and 4B models using 16, 16, and 32 A100 GPUs, with training times of approximately 4, 7, and 13 hours, respectively.
We will release the code, model checkpoints, and sampled training subset to facilitate reproducibility.

\noindent\textbf{Evaluation protocol.}
At test time, \textsc{SAVER} directly operates on sparsely sampled frames without using dense video inputs.
We control the visual budget by varying the sampling frame rate over
$\{0.1, 0.2, 0.5, 1, 2\}$ fps.
This protocol allows us to evaluate the performance--efficiency trade-off from extremely sparse observations to denser temporal inputs. We cap the number of sampled frames to $[4, 256]$ per video for fair comparison.
For each fps setting, we report both task performance and the corresponding average number of sampled frames.
We evaluate with \texttt{lmms-eval}~\citep{zhang2024lmmsevalrealitycheckevaluation} using greedy decoding unless noted.

\noindent\textbf{Evaluation benchmarks.}
Following prior video reasoning evaluation protocols~\citep{liu2026videoauto}, we evaluate \textsc{SAVER} on temporal grounding and video question answering benchmarks.
For temporal grounding, we evaluate on Charades-STA~\citep{gao2017tall} and ActivityNet~\citep{Heilbron_2015_CVPR}, reporting mean temporal intersection-over-union (mIoU).
We further evaluate on NExT-GQA~\citep{xiao2024can}, which combines question answering with temporal grounding and requires models to identify task-relevant video segments.
Beyond the grounding task used for post-training, we evaluate transfer to video-QA benchmarks.
For perception-oriented video-QA, we report accuracy on VideoMME without subtitles~\citep{fu2025video}, MVBench~\citep{li2024mvbench}, LongVideoBench~\citep{wu2024longvideobench}, and MMVU~\citep{zhao2025mmvu}.
For reasoning-intensive video understanding, we evaluate on VideoMMMU~\citep{hu2025video} and Minimal Video Pairs (MVP)~\citep{krojer2025shortcut}.

\begin{table*}[t]
\centering
\caption{
\textbf{Temporal grounding results under controlled frame budgets.}
For each Qwen3.5 setting, Qwen3.5 and \textsc{SAVER} are shown side by side under the same fps and average frame budget.
Numbers in parentheses indicate the absolute improvement over Qwen3.5 at the same fps.
}
\label{table:benchmark_grounding}
\scalebox{0.9}{
\begin{tabular}{lcc|lll|l}
\toprule
\textbf{Method}
& \textbf{FPS}
& \makecell{\textbf{Avg.}\\\textbf{Frames}}
& \textbf{Charades-STA}
& \textbf{ActivityNet}
& \textbf{NExT-GQA}
& \textbf{Avg. mIoU} \\
\midrule
\rowcolor{gray!12}
\multicolumn{7}{l}{\textit{\textbf{Prior temporal grounding models: 7B model}}} \\
Qwen2.5-VL-7B 
& 2.0 & 110 & \score{52.9} & \score{26.9} & \score{20.2} & \score{33.3} \\
TimeChat 
& 2.0 & 110 & \score{31.2} & \score{30.4} & \score{17.4} & \score{26.3} \\
Temporal-RLT
& 2.0 & 110 & \score{57.0} & \score{39.0} & \score{37.3} & \score{44.4} \\
VideoAuto-R1-7B
& 2.0 & 110 & \score{60.0} & \score{47.6} & \score{36.7} & \score{48.1} \\
\midrule

\rowcolor{gray!12}
\multicolumn{7}{l}{\textit{\textbf{Controlled frame-rate evaluation: 0.8B model}}} \\
Qwen3.5
& 2.0 & 110 & \score{43.9} & \score{38.0} & \score{26.6} & \score{36.1} \\
\rowcolor{blue!4}
\textbf{\textsc{SAVER}}
& 2.0 & 110
& \gain{50.5}{(+6.6)}
& \gain{45.3}{(+7.3)}
& \gain{28.4}{(+1.8)}
& \bgain{41.4}{(+5.3)} \\
Qwen3.5
& 0.5 & 31 & \score{33.4} & \score{28.4} & \score{19.0} & \score{26.9} \\
\rowcolor{blue!4}
\textbf{\textsc{SAVER}}
& 0.5 & 31
& \gain{40.7}{(+7.3)}
& \gain{42.4}{(+14.0)}
& \gain{17.6}{(-1.4)}
& \bgain{33.6}{(+6.7)} \\
Qwen3.5
& 0.1 & 7 & \score{24.0} & \score{21.7} & \score{10.3} & \score{18.7} \\
\rowcolor{blue!4}
\textbf{\textsc{SAVER}}
& 0.1 & 7
& \gain{31.1}{(+7.1)}
& \gain{32.7}{(+11.0)}
& \gain{15.6}{(+5.3)}
& \bgain{26.5}{(+7.8)} \\

\midrule
\rowcolor{gray!12}
\multicolumn{7}{l}{\textit{\textbf{Controlled frame-rate evaluation: 2B model}}} \\
Qwen3.5
& 2.0 & 110 & \score{44.0} & \score{38.3} & \score{22.6} & \score{34.9} \\
\rowcolor{blue!4}
\textbf{\textsc{SAVER}}
& 2.0 & 110
& \gain{54.3}{(+10.3)}
& \gain{44.8}{(+6.5)}
& \gain{37.1}{(+14.5)}
& \bgain{45.4}{(+10.5)} \\
Qwen3.5
& 0.5 & 31 & \score{32.3} & \score{29.8} & \score{21.7} & \score{27.9} \\
\rowcolor{blue!4}
\textbf{\textsc{SAVER}}
& 0.5 & 31
& \gain{45.2}{(+12.9)}
& \gain{43.0}{(+13.2)}
& \gain{38.9}{(+17.2)}
& \bgain{42.4}{(+14.5)} \\
Qwen3.5
& 0.1 & 7 & \score{17.6} & \score{20.7} & \score{16.9} & \score{18.4} \\
\rowcolor{blue!4}
\textbf{\textsc{SAVER}}
& 0.1 & 7
& \gain{28.7}{(+11.1)}
& \gain{37.2}{(+16.5)}
& \gain{24.2}{(+7.3)}
& \bgain{30.0}{(+11.6)} \\

\midrule
\rowcolor{gray!12}
\multicolumn{7}{l}{\textit{\textbf{Controlled frame-rate evaluation: 4B model}}} \\
Qwen3.5
& 2.0 & 110 & \score{54.2} & \score{44.7} & \score{8.1} & \score{35.7} \\
\rowcolor{blue!4}
\textbf{\textsc{SAVER}}
& 2.0 & 110
& \gain{54.8}{(+0.6)}
& \gain{46.9}{(+2.2)}
& \gain{38.2}{(+30.1)}
& \bgain{46.6}{(+10.9)} \\
Qwen3.5
& 0.5 & 31 & \score{43.9} & \score{40.5} & \score{3.9} & \score{29.4} \\
\rowcolor{blue!4}
\textbf{\textsc{SAVER}}
& 0.5 & 31
& \gain{48.0}{(+4.1)}
& \gain{43.4}{(+2.9)}
& \gain{31.3}{(+27.4)}
& \bgain{40.9}{(+11.5)} \\
Qwen3.5
& 0.1 & 7 & \score{25.0} & \score{29.8} & \score{4.3} & \score{19.7} \\
\rowcolor{blue!4}
\textbf{\textsc{SAVER}}
& 0.1 & 7
& \gain{29.5}{(+4.5)}
& \gain{32.5}{(+2.7)}
& \gain{22.7}{(+18.4)}
& \bgain{28.2}{(+8.5)} \\
\bottomrule
\end{tabular}
}
\vspace{-2mm}
\end{table*}

\begin{table*}[t]
\centering
\caption{
\textbf{Video-QA results under controlled frame budgets.}
\textsc{SAVER} consistently improves Qwen3.5 across model sizes and frame rates, matching or surpassing dense-frame performance with far fewer frames.
}
\label{table:benchmark_qa}
\scalebox{0.65}{
\begin{tabular}{lcc|llll|ll|l}
\toprule
\textbf{Method}
& \textbf{FPS}
& \makecell{\textbf{Avg.}\\\textbf{Frames}}
& \multicolumn{4}{c|}{\textbf{Video Perception}}
& \multicolumn{2}{c|}{\textbf{Video Reasoning}}
& \multirow{2}{*}{\textbf{\makecell{Avg.\\Acc.}}} \\
\cmidrule(lr){4-7}\cmidrule(lr){8-9}
& &
& \textbf{VideoMME}
& \textbf{MVBench}
& \textbf{LongVideoBench}
& \textbf{MMVU}
& \textbf{VideoMMMU}
& \textbf{MVP}
& \\
\midrule
\rowcolor{gray!12}
\multicolumn{10}{l}{\textit{\textbf{Prior video reasoning models (reported settings)}}} \\
Qwen2.5-VL-7B
& 2.0 & 139
& \score{66.0}
& \score{67.1}
& \score{60.9}
& \score{66.2}
& \score{54.7}
& \score{36.5}
& \score{58.6} \\
Qwen3-VL-8B
& 2.0 & 139
& \score{72.5}
& \score{69.4}
& \score{67.6}
& \score{69.9}
& \score{61.0}
& \score{40.5}
& \score{63.5} \\
VideoAuto-R1-7B
& 2.0 & 139
& \score{67.3}
& \score{71.0}
& \score{60.5}
& \score{69.7}
& \score{58.6}
& \score{39.4}
& \score{61.1} \\

\midrule
\rowcolor{gray!12}
\multicolumn{10}{l}{\textit{\textbf{Controlled frame-rate evaluation: 0.8B model}}} \\
Qwen3.5
& 2.0 & 139
& \score{55.0}
& \score{49.2}
& \score{54.2}
& \score{50.4}
& \score{32.7}
& \score{25.0}
& \score{44.4} \\
\rowcolor{blue!4}
\textbf{\textsc{SAVER}}
& 2.0 & 139
& \gain{56.3}{(+1.3)}
& \gain{52.9}{(+3.7)}
& \gain{54.7}{(+0.5)}
& \gain{51.2}{(+0.8)}
& \gain{40.8}{(+8.1)}
& \gain{23.4}{(-1.6)}
& \bgain{46.5}{(+2.1)} \\
Qwen3.5
& 0.5 & 97
& \score{54.0}
& \score{46.4}
& \score{52.1}
& \score{50.2}
& \score{35.3}
& \score{22.2}
& \score{43.4} \\
\rowcolor{blue!4}
\textbf{\textsc{SAVER}}
& 0.5 & 97
& \gain{55.0}{(+1.0)}
& \gain{49.6}{(+3.2)}
& \gain{52.3}{(+0.2)}
& \gain{47.9}{(-2.3)}
& \gain{39.1}{(+3.8)}
& \gain{21.2}{(-1.0)}
& \bgain{44.2}{(+0.8)} \\
Qwen3.5
& 0.1 & 38
& \score{49.0}
& \score{43.7}
& \score{49.1}
& \score{49.0}
& \score{31.6}
& \score{20.1}
& \score{40.4} \\
\rowcolor{blue!4}
\textbf{\textsc{SAVER}}
& 0.1 & 38
& \gain{50.3}{(+1.3)}
& \gain{46.0}{(+2.3)}
& \gain{49.5}{(+0.4)}
& \gain{50.1}{(+1.1)}
& \gain{38.3}{(+6.7)}
& \gain{17.6}{(-2.5)}
& \bgain{42.0}{(+1.6)} \\

\midrule
\rowcolor{gray!12}
\multicolumn{10}{l}{\textit{\textbf{Controlled frame-rate evaluation: 2B model}}} \\
Qwen3.5
& 2.0 & 139
& \score{62.6}
& \score{58.6}
& \score{55.9}
& \score{61.0}
& \score{38.8}
& \score{29.4}
& \score{51.0} \\
\rowcolor{blue!4}
\textbf{\textsc{SAVER}}
& 2.0 & 139
& \gain{61.1}{(-1.5)}
& \gain{56.9}{(-1.7)}
& \gain{60.5}{(+4.6)}
& \gain{61.8}{(+0.8)}
& \gain{57.0}{(+18.2)}
& \gain{30.5}{(+1.1)}
& \bgain{54.6}{(+3.6)} \\
Qwen3.5
& 0.5 & 97
& \score{61.7}
& \score{55.7}
& \score{52.4}
& \score{58.7}
& \score{40.2}
& \score{24.9}
& \score{48.9} \\
\rowcolor{blue!4}
\textbf{\textsc{SAVER}}
& 0.5 & 97
& \gain{59.9}{(-1.8)}
& \gain{57.0}{(+1.3)}
& \gain{59.0}{(+6.6)}
& \gain{60.2}{(+1.5)}
& \gain{57.9}{(+17.7)}
& \gain{26.5}{(+1.6)}
& \bgain{53.4}{(+4.5)} \\
Qwen3.5
& 0.1 & 38
& \score{55.9}
& \score{52.1}
& \score{51.3}
& \score{57.0}
& \score{37.4}
& \score{21.9}
& \score{45.9} \\
\rowcolor{blue!4}
\textbf{\textsc{SAVER}}
& 0.1 & 38
& \gain{53.6}{(-2.3)}
& \gain{54.3}{(+2.2)}
& \gain{56.0}{(+4.7)}
& \gain{59.7}{(+2.7)}
& \gain{56.1}{(+18.7)}
& \gain{24.9}{(+3.0)}
& \bgain{50.8}{(+4.9)} \\

\midrule
\rowcolor{gray!12}
\multicolumn{10}{l}{\textit{\textbf{Controlled frame-rate evaluation: 4B model}}} \\
Qwen3.5
& 2.0 & 139
& \score{70.9}
& \score{64.4}
& \score{64.7}
& \score{68.2}
& \score{54.8}
& \score{41.9}
& \score{60.8} \\
\rowcolor{blue!4}
\textbf{\textsc{SAVER}}
& 2.0 & 139
& \gain{71.8}{(+0.9)}
& \gain{65.3}{(+0.9)}
& \gain{65.6}{(+0.9)}
& \gain{70.9}{(+2.7)}
& \gain{69.8}{(+15.0)}
& \gain{43.1}{(+1.2)}
& \bgain{64.4}{(+3.6)} \\
Qwen3.5
& 0.5 & 97
& \score{69.5}
& \score{62.3}
& \score{62.6}
& \score{67.2}
& \score{54.2}
& \score{34.3}
& \score{58.4} \\
\rowcolor{blue!4}
\textbf{\textsc{SAVER}}
& 0.5 & 97
& \gain{70.3}{(+0.8)}
& \gain{61.4}{(-0.9)}
& \gain{63.8}{(+1.2)}
& \gain{67.7}{(+0.5)}
& \gain{71.2}{(+17.0)}
& \gain{34.6}{(+0.3)}
& \bgain{61.5}{(+3.1)} \\
Qwen3.5
& 0.1 & 38
& \score{63.2}
& \score{57.9}
& \score{59.3}
& \score{66.6}
& \score{52.4}
& \score{29.9}
& \score{54.9} \\
\rowcolor{blue!4}
\textbf{\textsc{SAVER}}
& 0.1 & 38
& \gain{64.0}{(+0.8)}
& \gain{59.1}{(+1.2)}
& \gain{60.0}{(+0.7)}
& \gain{68.0}{(+1.4)}
& \gain{72.7}{(+20.3)}
& \gain{30.8}{(+0.9)}
& \bgain{59.1}{(+4.2)} \\
\bottomrule
\end{tabular}
}
\end{table*}

\subsection{Main Results}

\noindent\textbf{Temporal grounding benchmarks.}
Table~\ref{table:benchmark_grounding} summarizes the results on Charades-STA, ActivityNet, and NExT-GQA.
We first evaluate temporal grounding because it is the task used for post-training and directly measures whether the model can localize task-relevant temporal evidence.
Temporal grounding is also highly sensitive to frame sparsity: accurate localization requires identifying event boundaries from limited temporal observations.
This is reflected in the original Qwen3.5 baselines. When the frame rate is reduced from 2.0 fps to 0.1 fps, the average mIoU of Qwen3.5-2B drops from 34.9\% to 18.4\%, confirming that sparse-frame temporal grounding is a challenging setting rather than a trivial reduction of input frames. \textsc{SAVER} substantially alleviates this degradation under the same frame budgets.
At 0.1 fps, \textsc{SAVER} improves average mIoU from 18.4\% to 30.0\% for the 2B model, yielding a large gain under extremely sparse observations.
The same trend also holds for the 4B model, where \textsc{SAVER} improves average mIoU from 19.7\% to 28.2\% at 0.1 fps.
Under denser inputs, \textsc{SAVER} remains effective, improving the 2B model from 34.9\% to 45.4\% at 2.0 fps.
These results show that dense-to-sparse reference learning helps the model recover task-relevant temporal evidence even when only a few frames are observed, while also improving temporal localization under denser views. The prior temporal grounding methods in the upper block are included as reported-reference results, often with different training data, model scales, and evaluation settings.
Our main comparison is the controlled frame-rate block, where Qwen3.5 and \textsc{SAVER} are evaluated under identical fps and frame budgets.
Under this controlled setting, \textsc{SAVER} consistently improves Qwen3.5 across model scales and frame rates.
This is notable because \textsc{SAVER} is trained with only 1,250 temporal grounding examples from Time-R1~\citep{wang2025time}, yet the learned sparse view policy generalizes across grounding benchmarks and frame budgets.
\begin{figure}[t]
    \centering
    \includegraphics[width=\linewidth]{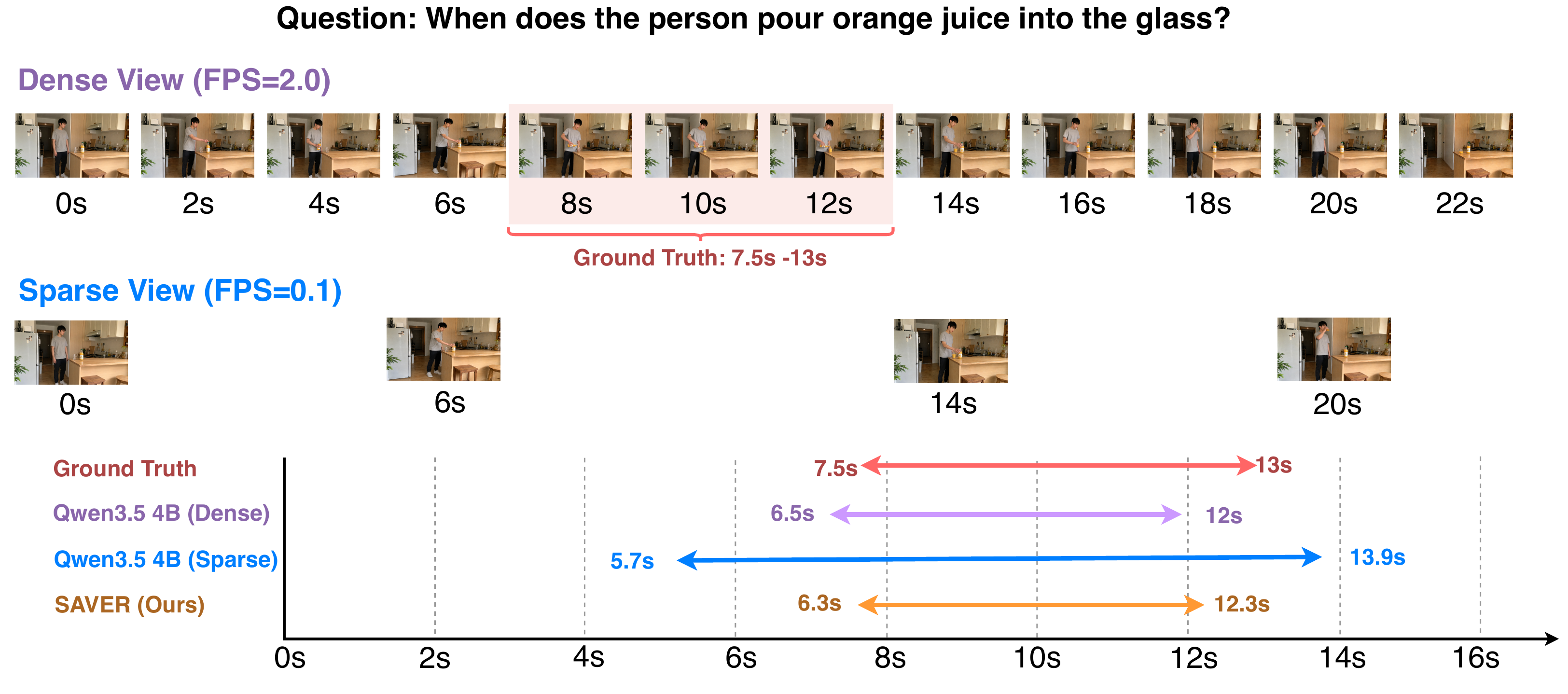}
\caption{
\textbf{Qualitative comparison under sparse observations.}
Dense frames provide finer temporal evidence, whereas sparse frames contain only a few sampled moments.
Under sparse input, Qwen3.5 predicts an overly broad interval, while \textsc{SAVER} produces a tighter prediction closer to the ground truth.
This shows that dense references help preserve task-relevant temporal evidence under sparse frame budgets.}
\label{fig:qualitative_sparse_grounding}
    \vspace{-2mm}
\end{figure}

\noindent\textbf{Transfer to video-QA benchmarks.}
We next evaluate transfer from grounding-only post-training to video question answering.
Table~\ref{table:benchmark_qa} reports results on six video-QA benchmarks, none of whose QA annotations are used during post-training.
Across the evaluated model scales and frame rates, \textsc{SAVER} improves macro-averaged QA accuracy over the corresponding Qwen3.5 baseline.
At 2.0 fps, average accuracy increases from 51.0\% to 54.6\% for the 2B model and from 60.8\% to 64.4\% for the 4B model.
At 0.1 fps, the 2B model improves from 45.9\% to 50.8\%.
The gains vary substantially across benchmarks.
VideoMMMU shows particularly large improvements, including an increase from 52.4\% to 72.7\% for the 4B model at 0.1 fps.
Some settings show regressions, such as VideoMME for the 2B model and MVP for the 0.8B model.
The results therefore support improved aggregate video-QA performance following grounding-only post-training, while also revealing task-dependent transfer.
Sparse inference also reduces the visual input budget.
Across the six video-QA benchmarks, changing the sampling rate from 2.0 to 0.1 fps reduces the average number of observed frames from 139 to 38, a 72.7\% reduction.
Separately, profiling the original Qwen3.5-2B model under the settings in Appendix~\ref{app:efficiency_profiling} reports reductions of 86.8\% in visual tokens, 49.3\% in peak GPU memory, and 63.6\% in end-to-end latency, averaged across nine benchmarks.
These measurements characterize the cost of the profiled input settings on the base model; they do not establish a \textsc{SAVER}-specific speedup.

\noindent\textbf{Qualitative analysis.}
Figure~\ref{fig:qualitative_sparse_grounding} provides a qualitative example of sparse-frame temporal grounding.
When only a few frames are available, the original Qwen3.5 model captures the rough event context but predicts a much broader temporal interval, reflecting uncertainty caused by missing intermediate frames.
In contrast, \textsc{SAVER} produces a more precise segment that is close to both the ground-truth interval and the dense view prediction.
This suggests that the proposed dense-to-sparse alignment does not simply improve numerical scores, but also helps the model retain task-relevant temporal evidence when operating under sparse observations.

\subsection{Analysis}

\noindent\textbf{Reward ablation.}
Table~\ref{tab:ablation_rewards} reports reward ablations using Qwen3.5-2B evaluated at 0.1 fps.
Joint dense-and-sparse training obtains 27.4 mIoU and 48.7\% average video-QA accuracy, exceeding either single-view variant.
Adding the reference reward increases these scores to 29.1 and 49.3\%, respectively.
The gated variant obtains the highest scores in this comparison, reaching 30.0 mIoU and 50.8\% average video-QA accuracy.
These ablations show incremental gains from the reference reward and its relative-quality gate under the evaluated configuration.

\noindent\textbf{Comparison with alternative training strategies.}

We compare alternative post-training strategies in Table~\ref{tab:alternative_training}.
All variants use the same Qwen3.5-2B initialization and 1,250 grounding examples and are evaluated at 0.1 fps.
Time-R1 RL uses 2.0-fps training inputs and obtains 21.0 mIoU and 49.1\% average video-QA accuracy.
Sparse SFT uses 0.5-fps inputs with either ground-truth segments or fixed pseudo-labels generated by the original model at 2.0 fps, obtaining 20.3 and 19.6 mIoU, respectively.
\textsc{SAVER} achieves the highest scores among these evaluated configurations, with 30.0 mIoU and 50.8\% average video-QA accuracy.
Replacing the dense reference with the ground-truth segment gives 29.5 mIoU and 50.1\% average video-QA accuracy.
The two reference variants perform similarly on grounding, while the dense-reference variant obtains a modestly higher QA score in this comparison.
These results support the effectiveness of the proposed training configuration; the mechanism underlying its cross-task transfer remains an open question.

\begin{table}[H]

\centering

\begin{minipage}[t]{0.52\textwidth}
\centering
\caption{
\textbf{Reward ablation.}
}

\label{tab:ablation_rewards}

\scalebox{0.78}{
\begin{tabular}{lcccc|cc}
\toprule
\textbf{Variant}
& $R_{\mathrm{fmt}}$
& $R_{\mathrm{grd}}$
& $R_{\mathrm{ref}}$
& $w^d$
& \makecell{TG\\mIoU}
& \makecell{QA\\Avg.} \\
\midrule

Qwen3.5
& -- & -- & -- & -- 
& 18.4 & 45.9 \\

Sparse-only
& \checkmark & \checkmark & -- & --
& 19.0 & 48.5 \\ 

Dense-only
& \checkmark & \checkmark & -- & -- 
& 24.2 & 48.3 \\

Dense+Sparse
& \checkmark & \checkmark & -- & -- 
& 27.4 & 48.7 \\

+ Ref. reward
& \checkmark & \checkmark & \checkmark & -- 
& 29.1 & 49.3 \\

\rowcolor{blue!4}
\textsc{SAVER}
& \checkmark & \checkmark & \checkmark & \checkmark 
& \textbf{30.0} & \textbf{50.8} \\

\bottomrule
\end{tabular}
}
\end{minipage}
\hfill
\begin{minipage}[t]{0.457\textwidth}
\centering
\caption{
\textbf{Alternative training strategies.}
}

\label{tab:alternative_training}

\scalebox{0.75}{
\begin{tabular}{l|cc}
\toprule
\textbf{Method}
& \makecell{TG\\mIoU}
& \makecell{QA\\Avg.} \\
\midrule

Qwen3.5
& 18.4 & 45.9 \\

Time-R1 RL
& 21.0 & 49.1 \\

Sparse SFT (GT labels)
& 20.3 & 48.1 \\

Sparse SFT (dense pseudo-labels)
& 19.6 & 49.5 \\

SAVER(GT ref.)    
& 29.5 & 50.1 \\

\rowcolor{blue!4}
\textsc{SAVER}(dense ref.)    
& \textbf{30.0} & \textbf{50.8} \\

\bottomrule
\end{tabular}
}

\end{minipage}

\end{table}

\begin{wrapfigure}{r}{0.55\textwidth}
    \vspace{-5mm}
    \centering
    \includegraphics[width=0.45\textwidth]{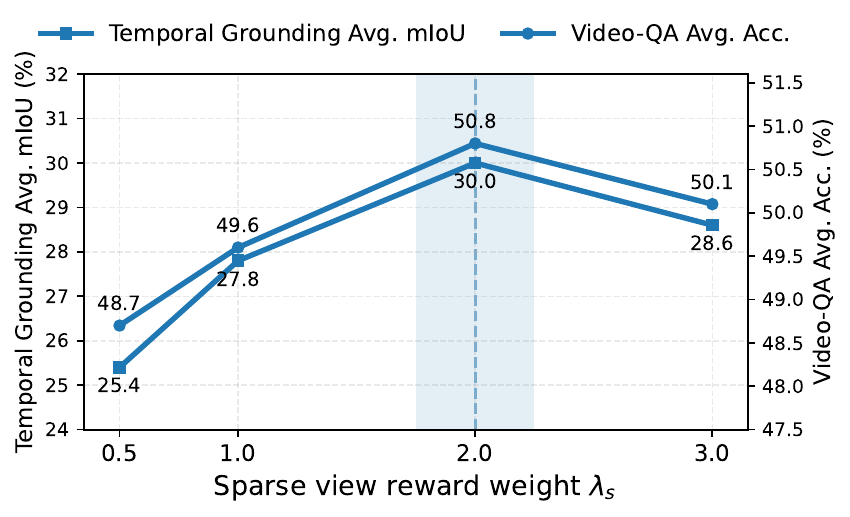}
        \vspace{-3mm}
    \caption{
    \textbf{Effect of sparse view reward weight.}
    }
    \label{fig:ablation_sparse_weight}
    \vspace{-5mm}
\end{wrapfigure}

\noindent\textbf{Effect of sparse view reward weight.}
Figure~\ref{fig:ablation_sparse_weight} shows the effect of the sparse view reward weight $\lambda_s$ at 0.1 fps, where sparse-frame inference is most challenging.
A small value of $\lambda_s$ 
provides insufficient
pressure for the model to improve sparse-frame robustness, leading to weaker performance under limited frame budgets.
Increasing $\lambda_s$ improves both sparse view 
temporal grounding mIoU and video-QA accuracy, showing that explicitly emphasizing the sparse view is important for sparse video 
reasoning.
However, an overly large weight can overemphasize sparse view rewards and weaken the balance between dense view scaffolding and sparse view optimization.
We therefore set $\lambda_s=2$ as the default, which achieves the best overall trade-off across temporal grounding and video-QA. We further analyze the sensitivity to the reference reward weight
$\lambda_{\mathrm{ref}}$ in Appendix~\ref{app:ref_reward_weight},
where the results show an optimal intermediate weighting rather than
monotonic gains from increasing the reward magnitude.

\par\WFclear
\paragraph{Limitations and failure cases.}
While \textsc{SAVER} improves sparse-frame inference across benchmarks, it does not eliminate all failures under limited temporal evidence.
We observe two common failure modes.
First, when the sparse frames completely miss the key event, the model may still predict an overly broad or shifted temporal interval.
Second, when visually similar actions occur multiple times in a video, the model may align to the wrong occurrence, especially under very low frame rates.
We provide additional success and failure cases in the Appendix~\ref{app:failure_cases}. A promising future direction is adaptive online sampling, where the model can request additional frames when sparse observations are insufficient.

\section{Conclusion}

We presented \textsc{SAVER}, a dense-to-sparse reinforcement post-training framework for sparse-input video understanding.
The framework jointly trains dense and sparse views with grounding supervision and uses a relative-quality gate to control an interval-level reference reward.
Using 1,250 grounding examples and no video-QA annotations during post-training, \textsc{SAVER} improves macro-averaged performance across the evaluated model scales and frame budgets.
Its video-QA results demonstrate cross-task transfer, although the magnitude and direction of the gains vary across benchmarks. Importantly, inference requires neither dense video pre-scanning nor adaptive frame selection or test-time optimization, preserving a fixed sparse-input protocol. Rather than selecting more informative frames at inference time, \textsc{SAVER} learns to make better use of limited temporal evidence through post-training.
These findings support training-time dense references as a practical approach to improving performance under limited temporal observations.
\section*{Acknowledgments}
\noindent This work was supported by the European Union’s Horizon Europe research and innovation programme under grant agreement number 101214398 (ELLIOT).

\section*{AI use statement}
In this work, we used generative AI tools to assist with developing conceptual frameworks, refining research hypotheses, discussing reward formulations and experimental designs, implementing and debugging code, reviewing literature, preparing figures, and drafting and editing the manuscript. We did not use AI to generate synthetic datasets, develop theoretical models, formulate mathematical claims, contribute to mathematical proofs, assist with translation, clean or reformat datasets, support qualitative or thematic data analysis, or interpret experimental results. All AI-assisted work was critically reviewed and verified by the authors. AI-assisted code was manually tested, figures and tables were checked against original experimental results, and literature references were verified against source publications. The authors take full responsibility for the methodology, results, claims, and final content of this work.

\subsection*{Reproducibility statement}

We provide the training and evaluation protocols,
implementation details, and computational resources
in Section 4 and the Appendix. We plan to release
our code, model checkpoints, and sampled training
subset to facilitate reproducibility.

\newpage

\bibliography{iclr2027_conference}
\bibliographystyle{iclr2027_conference}

\appendix
\newpage
\section{Controlled Frame-Rate Evaluation}
We provide detailed results under the controlled frame-rate protocol in
Table~\ref{tab:appendix_temporal_grounding} and Table~\ref{tab:appendix_video_qa}.
For each fps setting, we report both the task performance and the corresponding average number of observed frames for each benchmark individually.
To ensure stable and meaningful video evaluation, the total number of sampled frames is capped within $[4,256]$ per video.
We additionally report the average frame usage both for each individual benchmark and across all temporal grounding or video-QA benchmarks. The detailed results further demonstrate that reducing the sampling rate substantially lowers the visual input budget across all benchmarks, but also makes sparse-frame reasoning significantly more challenging for the original Qwen3.5 models.
In contrast, \textsc{SAVER} improves average performance under matched frame budgets across all model scales and fps settings, for both temporal grounding and video-QA.
In most settings, the largest gains appear at 0.5\,fps or below, where temporal evidence becomes highly sparse and incomplete. Notably, \textsc{SAVER} substantially narrows the sparse--dense performance gap, and in several cases sparse \textsc{SAVER} inference approaches or even surpasses denser Qwen3.5 baselines while using far fewer observed frames.
These results suggest that dense-to-sparse post-training improves the robustness and reliability of video reasoning under limited temporal evidence, rather than merely reducing the number of processed frames.

\begin{table*}[h]
    \centering
    \footnotesize
    \setlength{\tabcolsep}{0.75em}
\caption{\textbf{Temporal grounding results under controlled frame budgets.} For each fps setting, we report grounding performance and the corresponding number of observed frames for three model scales (0.8B, 2B, and 4B). Avg. Frames denotes the average number of observed frames across all grounding benchmarks, and Avg. mIoU denotes the average grounding performance.}
   \label{tab:appendix_temporal_grounding}
    \resizebox{\textwidth}{!}{
    \begin{tabular}{lccc|cccccc|c}
    \toprule
    \multirow{2}{*}{\textbf{Model}}
    & \multirow{2}{*}{\textbf{Size}}
    & \multirow{2}{*}{\textbf{FPS}}
    & \multirow{2}{*}{\textbf{\makecell{Avg.\\Frames}}}
    & \multicolumn{6}{c|}{\textbf{Temporal Grounding}}
    & \multirow{2}{*}{\textbf{\makecell{Avg.\\mIoU}}} \\
    \cmidrule(lr){5-10}
    & & &
    & Charades-STA
    & \makecell{Frames}
    & ActivityNet
    & \makecell{Frames}
    & NExT-GQA
    & \makecell{Frames}
    & \\
    \midrule

    \rowcolor{gray!12}
    \multicolumn{11}{l}{\textit{\textbf{Controlled frame-rate evaluation: 0.8B model}}}\\

    Qwen3.5
    & 0.8B & 2.0 & 110
    & 43.9 & 59
    & 38.0 & 191
    & 26.6 & 80
    & 36.1 \\

    Qwen3.5
    & 0.8B & 1.0 & 63
    & 39.1 & 29
    &  34.4 & 121
    & 24.0 & 39
    & 32.5 \\

    Qwen3.5
    & 0.8B & 0.5 & 31
    & 33.4 & 14
    & 28.4 & 60
    & 19.0 & 19
    & 26.9 \\

    Qwen3.5
    & 0.8B & 0.2 & 12
    & 24.4 & 5
    & 23.1 & 24
    & 12.5 & 8
    & 20.0 \\

    Qwen3.5
    & 0.8B & 0.1 & 7
    & 24.0 & 4
    & 21.7 & 12
    & 10.3 & 5
    & 18.7 \\

    \rowcolor{blue!4}
    \textbf{\textsc{SAVER}}
    & \textbf{0.8B} & \textbf{2.0} & \textbf{110}
    & 50.5 & 59
    & 45.3 & 191
    & 28.4 & 80
    & \textbf{41.4} \\

    \rowcolor{blue!4}
    \textbf{\textsc{SAVER}}
    & \textbf{0.8B} & \textbf{1.0} & \textbf{63}
    &  47.2& 29
    & 44.3 & 121
    & 15.6  & 39
    & \textbf{35.7} \\

    \rowcolor{blue!4}
    \textbf{\textsc{SAVER}}
    & \textbf{0.8B} & \textbf{0.5} & \textbf{31}
    & 40.7 & 14
    & 42.4 & 60
    & 17.6 & 19
    & \textbf{33.6} \\

    \rowcolor{blue!4}
    \textbf{\textsc{SAVER}}
    & \textbf{0.8B} & \textbf{0.2} & \textbf{12}
    & 31.5& 5
    & 38.8 & 24
    & 18.0 & 8
    &  \textbf{29.4} \\

    \rowcolor{blue!4}
    \textbf{\textsc{SAVER}}
    & \textbf{0.8B} & \textbf{0.1} & \textbf{7}
    & 31.1 & 4
    & 32.7 & 12
    & 15.6 & 5
    & \textbf{26.5} \\

    \midrule

    \rowcolor{gray!12}
    \multicolumn{11}{l}{\textit{\textbf{Controlled frame-rate evaluation: 2B model}}}\\

    Qwen3.5
    & 2B & 2.0 & 110
    & 44.0 & 59
    & 38.3 & 191
    & 22.6 & 80
    & 34.9 \\

    Qwen3.5
    & 2B & 1.0 & 63
    & 39.9 & 29
    &  34.5 & 121
    & 24.6 & 39
    & 33.0 \\

    Qwen3.5
    & 2B & 0.5 & 31
    & 32.3 & 14
    & 29.8 & 60
    & 21.7 & 19
    & 27.9 \\

    Qwen3.5
    & 2B & 0.2 & 12
    & 17.9 & 5
    & 22.6 & 24
    & 18.2 & 8
    & 19.5 \\

    Qwen3.5
    & 2B & 0.1 & 7
    & 17.6 & 4
    & 20.7 & 12
    & 16.9 & 5
    & 18.4 \\

    \rowcolor{blue!4}
    \textbf{\textsc{SAVER}}
    & \textbf{2B} & \textbf{2.0} & \textbf{110}
    & 54.3 & 59
    & 44.8 & 191
    & 37.1 & 80
    & \textbf{45.4} \\

    \rowcolor{blue!4}
    \textbf{\textsc{SAVER}}
    & \textbf{2B} & \textbf{1.0} & \textbf{63}
    & 52.0 & 29
    & 44.2 & 121
    & 34.9 & 39
    & \textbf{43.7} \\

    \rowcolor{blue!4}
    \textbf{\textsc{SAVER}}
    & \textbf{2B} & \textbf{0.5} & \textbf{31}
    & 45.2 & 14
    & 43.0 & 60
    & 38.9 & 19
    & \textbf{42.4} \\

    \rowcolor{blue!4}
    \textbf{\textsc{SAVER}}
    & \textbf{2B} & \textbf{0.2} & \textbf{12}
    &31.2  & 5
    & 40.4 & 24
    & 25.5 & 8
    & \textbf{32.3} \\

    \rowcolor{blue!4}
    \textbf{\textsc{SAVER}}
    & \textbf{2B} & \textbf{0.1} & \textbf{7}
    & 28.7 & 4
    & 37.2 & 12
    & 24.2 & 5
    & \textbf{30.0} \\

    \midrule

    \rowcolor{gray!12}
    \multicolumn{11}{l}{\textit{\textbf{Controlled frame-rate evaluation: 4B model}}}\\

    Qwen3.5
    & 4B & 2.0 & 110
    & 54.2 & 59
    & 44.7 & 191
    & 8.1 & 80
    & 35.7 \\

    Qwen3.5
    & 4B & 1.0 & 63
    &  51.5& 29
    & 43.5 & 121
    & 6.3 & 39
    &33.8 \\

    Qwen3.5
    & 4B & 0.5 & 31
    & 43.9 & 14
    & 40.5 & 60
    & 3.9 & 19
    & 29.4 \\

    Qwen3.5
    & 4B & 0.2 & 12
    & 29.3 & 5
    & 36.4 & 24
    & 4.1 & 8
    & 23.2 \\

    Qwen3.5
    & 4B & 0.1 & 7
    & 25.0 & 4
    & 29.8 & 12
    & 4.3 & 5
    & 19.7 \\

    \rowcolor{blue!4}
    \textbf{\textsc{SAVER}}
    & \textbf{4B} & \textbf{2.0} & \textbf{110}
    & \textbf{54.8} & 59
    & \textbf{46.9} & 191
    & \textbf{38.2} & 80
    & \textbf{46.6} \\

    \rowcolor{blue!4}
    \textbf{\textsc{SAVER}}
    & \textbf{4B} & \textbf{1.0} & \textbf{63}
    & 53.4 & 29
    & 46.4 & 121
    & 34.5 & 39
    & \textbf{44.7} \\

    \rowcolor{blue!4}
    \textbf{\textsc{SAVER}}
    & \textbf{4B} & \textbf{0.5} & \textbf{31}
    & 48.0 & 14
    & 43.4 & 60
    & 31.3 & 19
    & \textbf{40.9} \\

    \rowcolor{blue!4}
    \textbf{\textsc{SAVER}}
    & \textbf{4B} & \textbf{0.2} & \textbf{12}
    & 33.8 & 5
    & 38.4 & 24
    & 25.0 & 8
    & \textbf{32.4} \\

    \rowcolor{blue!4}
    \textbf{\textsc{SAVER}}
    & \textbf{4B} & \textbf{0.1} & \textbf{7}
    & 29.5 & 4
    & 32.5 & 12
    & 22.7 & 5
    & \textbf{28.2} \\

    \bottomrule
    \end{tabular}
    }
    \vspace{-5mm}
\end{table*}

\begin{table*}[h]
    \centering
    \footnotesize
    \setlength{\tabcolsep}{0.35em}
\caption{
\textbf{Video-QA results under controlled frame budgets.} For each fps setting, we report benchmark performance and the corresponding number of observed frames across video perception and video reasoning tasks for three model scales (0.8B, 2B, and 4B). Avg. Frames denotes the average number of observed frames across all benchmarks, and Avg. Acc. denotes the average performance.
}
\label{tab:appendix_video_qa}
    \resizebox{\textwidth}{!}{
    \begin{tabular}{lccc|cccccccc|cccc|c}
    \toprule
    \multirow{2}{*}{\textbf{Model}}
    & \multirow{2}{*}{\textbf{Size}}
    & \multirow{2}{*}{\textbf{FPS}}
    & \multirow{2}{*}{\textbf{\makecell{Avg.\\Frames}}}
    & \multicolumn{8}{c|}{\textbf{Video Perception}}
    & \multicolumn{4}{c|}{\textbf{Video Reasoning}}
    & \multirow{2}{*}{\textbf{\makecell{Avg.\\Acc.}}} \\
    \cmidrule(lr){5-12}\cmidrule(lr){13-16}
    & & &
    & VideoMME & \makecell{Frames}
    & MVBench & \makecell{Frames}
    & LongVideoBench & \makecell{Frames}
    & MMVU & \makecell{Frames}
    & VideoMMMU & \makecell{Frames}
    & MVP & \makecell{Frames}
    & \\
    \midrule

    \rowcolor{gray!12}
    \multicolumn{17}{l}{\textit{\textbf{Controlled frame-rate evaluation: 0.8B model}}}\\

    Qwen3.5
    & 0.8B & 2.0 & 139
    & 55.0 & 224
    & 49.2 & 35
    & 54.2 & 200
    & 50.4 & 99
    & 32.7 & 253
    & 25.0 & 25
    & 44.4 \\

    Qwen3.5
    & 0.8B & 1.0 & 118
    & 54.8 & 197
    & 47.6 & 17
    & 52.8 & 190
    & 50.1 & 51
    &  33.1& 240
    & 24.7 & 12
    & 43.9 \\

    Qwen3.5
    & 0.8B & 0.5 & 97
    & 54.0 & 171
    & 46.4 & 9
    & 52.1 & 170
    & 50.2 & 25
    & 35.3 & 200
    & 22.2 & 6
    & 43.4 \\

    Qwen3.5
    & 0.8B & 0.2 & 61
    & 51.6 & 124
    & 44.3 & 5
    & 50.9 & 122
    & 49.4 & 10
    & 34.1 & 99
    &  21.1 & 5
    & 41.9 \\

    Qwen3.5
    & 0.8B & 0.1 & 38
    & 49.0 & 96
    & 43.7 & 4
    & 49.1 & 71
    & 49.0 & 6
    & 31.6 & 50
    & 20.1 & 4
    & 40.4 \\

    \rowcolor{blue!4}
    \textbf{\textsc{SAVER}}
    & \textbf{0.8B} & \textbf{2.0} & \textbf{139}
    & 56.3 & 224
    & 52.9 & 35
    & 54.7 & 200
    & 51.2 & 99
    & 40.8 & 253
    & 23.4 & 25
    & \textbf{46.5} \\

    \rowcolor{blue!4}
    \textbf{\textsc{SAVER}}
    & \textbf{0.8B} & \textbf{1.0} & \textbf{118}
    & 56.2 & 197
    & 51.1 & 17
    & 53.3 & 190
    & 50.6 & 51
    & 38.9 & 240
    & 22.3 & 12
    & \textbf{45.4} \\

    \rowcolor{blue!4}
    \textbf{\textsc{SAVER}}
    & \textbf{0.8B} & \textbf{0.5} & \textbf{97}
    & 55.0 & 171
    & 49.6 & 9
    & 52.3 & 170
    & 47.9 & 25
    & 39.1 & 200
    & 21.2 & 6
    & \textbf{44.2} \\

    \rowcolor{blue!4}
    \textbf{\textsc{SAVER}}
    & \textbf{0.8B} & \textbf{0.2} & \textbf{61}
    & 52.5 & 124
    & 47.1 & 5
    & 50.5 & 122
    & 45.8 & 10
    & 39.7 & 99
    & 19.9 & 5
    & \textbf{42.6}\\

    \rowcolor{blue!4}
    \textbf{\textsc{SAVER}}
    & \textbf{0.8B} & \textbf{0.1} & \textbf{38}
    & 50.3 & 96
    & 46.0 & 4
    & 49.5 & 71
    & 50.1 & 6
    & 38.3 & 50
    & 17.6 & 4
    & \textbf{42.0} \\

    \midrule

    \rowcolor{gray!12}
    \multicolumn{17}{l}{\textit{\textbf{Controlled frame-rate evaluation: 2B model}}}\\

    Qwen3.5
    & 2B & 2.0 & 139
    & 62.6 & 224
    & 58.6 & 35
    & 55.9 & 200
    & 61.0 & 99
    & 38.8 & 253
    & 29.4 & 25
    & 51.0 \\

    Qwen3.5
    & 2B & 1.0 & 118
    & 62.4 & 197
    & 56.2 & 17
    & 54.5& 190
    & 60.0 & 51
    & 38.3 & 240
    & 27.4 & 12
    & 49.8\\

    Qwen3.5
    & 2B & 0.5 & 97
    & 61.7 & 171
    & 55.7 & 9
    & 52.4 & 170
    & 58.7 & 25
    & 40.2 & 200
    & 24.9 & 6
    & 48.9 \\

    Qwen3.5
    & 2B & 0.2 & 61
    & 59.3 & 124
    & 53.3 & 5
    & 51.9 & 122
    & 57.8 & 10
    & 40.3 & 99
    & 22.8 & 5
    & 47.6 \\

    Qwen3.5
    & 2B & 0.1 & 38
    & 55.9 & 96
    & 52.1 & 4
    & 51.3 & 71
    & 57.0 & 6
    & 37.4 & 50
    & 21.9 & 4
    & 45.9 \\

    \rowcolor{blue!4}
    \textbf{\textsc{SAVER}}
    & \textbf{2B} & \textbf{2.0} & \textbf{139}
    & 61.1 & 224
    & 56.9 & 35
    & 60.5 & 200
    & 61.8 & 99
    & 57.0 & 253
    & 30.5 & 25
    & \textbf{54.6} \\

    \rowcolor{blue!4}
    \textbf{\textsc{SAVER}}
    & \textbf{2B} & \textbf{1.0} & \textbf{118}
    & 61.0 & 197
    & 56.6 & 17
    & 59.8 & 190
    & 61.4 & 51
    & 56.0 & 240
    & 28.3 & 12
    & \textbf{53.9} \\

    \rowcolor{blue!4}
    \textbf{\textsc{SAVER}}
    & \textbf{2B} & \textbf{0.5} & \textbf{97}
    & 59.9 & 171
    & 57.0 & 9
    & 59.0 & 170
    & 60.2 & 25
    & 57.9 & 200
    & 26.5 & 6
    & \textbf{53.4} \\

    \rowcolor{blue!4}
    \textbf{\textsc{SAVER}}
    & \textbf{2B} & \textbf{0.2} & \textbf{61}
    & 56.9 & 124
    & 54.5 & 5
    & 56.8 & 122
    & 60.6 & 10
    & 57.7 & 99
    & 24.5 & 5
    & \textbf{51.8} \\

    \rowcolor{blue!4}
    \textbf{\textsc{SAVER}}
    & \textbf{2B} & \textbf{0.1} & \textbf{38}
    & 53.6 & 96
    & 54.3 & 4
    & 56.0 & 71
    & 59.7 & 6
    & 56.1 & 50
    & 24.9 & 4
    & \textbf{50.8} \\

    \midrule

    \rowcolor{gray!12}
    \multicolumn{17}{l}{\textit{\textbf{Controlled frame-rate evaluation: 4B model}}}\\

    Qwen3.5
    & 4B & 2.0 & 139
    & 70.9 & 224
    & 64.4 & 35
    & 64.7 & 200
    & 68.2 & 99
    & 54.8 & 253
    & 41.9 & 25
    & 60.8 \\

    Qwen3.5
    & 4B & 1.0 & 118
    & 70.1 & 197
    & 62.6 & 17
    & 64.3 & 190
    & 68.7 & 51
    & 54.8 & 240
    & 36.5 & 12
    & 59.5 \\

    Qwen3.5
    & 4B & 0.5 & 97
    & 69.5 & 171
    & 62.3 & 9
    & 62.6 & 170
    & 67.2 & 25
    & 54.2 & 200
    & 34.3 & 6
    & 58.4 \\

    Qwen3.5
    & 4B & 0.2 & 61
    & 66.9 & 124
    & 58.7 & 5
    & 59.8 & 122
    & 67.4  & 10
    &  54.1 & 99
    & 30.3  & 5
    & 56.2  \\

    Qwen3.5
    & 4B & 0.1 & 38
    & 63.2 & 96
    & 57.9 & 4
    & 59.3 & 71
    & 66.6 & 6
    & 52.4 & 50
    & 29.9 & 4
    & 54.9 \\

    \rowcolor{blue!4}
    \textbf{\textsc{SAVER}}
    & \textbf{4B} & \textbf{2.0} & \textbf{139}
    & \textbf{71.8} & 224
    & \textbf{65.3} & 35
    & \textbf{65.6} & 200
    & \textbf{70.9} & 99
    & 69.8 & 253
    & \textbf{43.1} & 25
    & \textbf{64.4} \\

    \rowcolor{blue!4}
    \textbf{\textsc{SAVER}}
    & \textbf{4B} & \textbf{1.0} & \textbf{118}
    & 71.3 & 197
    & 62.4 & 17
    & 64.5 & 190
    & 70.2 & 51
    & 71.1 & 240
    & 36.5 & 12
    & \textbf{62.7} \\

    \rowcolor{blue!4}
    \textbf{\textsc{SAVER}}
    & \textbf{4B} & \textbf{0.5} & \textbf{97}
    & 70.3 & 171
    & 61.4 & 9
    & 63.8 & 170
    & 67.7 & 25
    & 71.2 & 200
    & 34.6 & 6
    & \textbf{61.5} \\

    \rowcolor{blue!4}
    \textbf{\textsc{SAVER}}
    & \textbf{4B} & \textbf{0.2} & \textbf{61}
    & 67.0 & 124
    & 59.6 & 5
    & 62.0 & 122
    & 67.8 & 10
    & 72.6 & 99
    & 30.6 & 5
    &\textbf{60.0} \\

    \rowcolor{blue!4}
    \textbf{\textsc{SAVER}}
    & \textbf{4B} & \textbf{0.1} & \textbf{38}
    & 64.0 & 96
    & 59.1 & 4
    & 60.0 & 71
    & 68.0 & 6
    & \textbf{72.7} & 50
    & 30.8 & 4
    & \textbf{59.1} \\

    \bottomrule
    \end{tabular}
    }
    \vspace{-2mm}
\end{table*}

\section{Effect of sparse view Training Frame Rate}

Table~\ref{tab:sparse_train_fps} studies the effect of the sparse view frame rate used during post-training, while fixing the dense view frame rate to 2.0 fps. We evaluate all variants under 0.1-fps inference, where sparse-frame reasoning is most challenging. Using an excessively sparse training view (0.1 fps) weakens the temporal evidence available to the sparse branch during training, making dense-to-sparse alignment less effective. In contrast, using a less sparse training view (1.0 fps) reduces the discrepancy between dense and sparse observations, weakening the robustness gains under sparse inference. We find that a moderate sparse view frame rate (0.5 fps) provides the best trade-off, achieving the strongest performance on both temporal grounding and video-QA benchmarks.

\begin{table}[t]
\centering
\small
\setlength{\tabcolsep}{6pt}
\caption{
\textbf{Effect of sparse view training frame rate.}
We fix the dense view training frame rate to 2.0 fps and vary the sparse view frame rate used during post-training.
Results are evaluated under 0.1-fps inference, where sparse-frame reasoning is most challenging.
}
\vspace{2mm}

\label{tab:sparse_train_fps}
\begin{tabular}{lcccc}
\toprule
\textbf{Model}
& \textbf{Size}
& \textbf{Sparse View FPS}
& \textbf{TG Avg. mIoU}
& \textbf{Video-QA Avg. Acc.} \\
\midrule
\textsc{SAVER}
& 2B
& 0.1
& 29.6
& 49.7 \\
\rowcolor{blue!4}
\textsc{SAVER}
& 2B
& \textbf{0.5}
& \textbf{30.0}
& \textbf{50.8} \\

\textsc{SAVER}
& 2B
& 1.0
& 28.4
& 48.8 \\
\bottomrule
\end{tabular}
\end{table}

\section{Practical Efficiency Profiling}
\label{app:efficiency_profiling}

To complement the sampled-frame analysis in the main paper, we further profile
the practical inference cost of sparse video processing in terms of visual-token
count, peak GPU memory, end-to-end latency, GPU inference latency, and estimated
computation.
We evaluate Qwen3.5-2B at 2.0 and 0.1 fps on the same three temporal-grounding
and six video-QA benchmarks used in the main experiments.
All measurements are collected on a single 96-GB NVIDIA H100 GPU with batch
size 1 and BF16 precision.
We perform one warm-up generation before measurement and use CUDA
synchronization for latency profiling.
Table~\ref{tab:efficiency_profiling} reports each metric as 2.0 fps / 0.1 fps.

\begin{table*}[t]
\centering
\caption{
\textbf{Practical efficiency profiling of Qwen3.5-2B at 2.0 and 0.1 fps.}
Lower frame rates substantially reduce visual-token count,
peak GPU memory, and inference latency on average across
nine evaluation benchmarks.
}
\label{tab:efficiency_profiling}
\vspace{-2mm}

\resizebox{\textwidth}{!}{
\begin{tabular}{l*{8}{>{\centering\arraybackslash}p{1.5cm}}}
\toprule
& \multicolumn{2}{c}{\textbf{Visual tokens} $\downarrow$}
& \multicolumn{2}{c}{\textbf{Peak memory (GiB)} $\downarrow$}
& \multicolumn{2}{c}{\textbf{E2E latency (s)} $\downarrow$}
& \multicolumn{2}{c}{\textbf{GPU inference (s)} $\downarrow$} \\
\cmidrule(lr){2-3}
\cmidrule(lr){4-5}
\cmidrule(lr){6-7}
\cmidrule(lr){8-9}
\textbf{Benchmark}
& \textbf{2.0 fps} & \textbf{0.1 fps}
& \textbf{2.0 fps} & \textbf{0.1 fps}
& \textbf{2.0 fps} & \textbf{0.1 fps}
& \textbf{2.0 fps} & \textbf{0.1 fps} \\
\midrule

Charades-STA
& 3,746 & 251
& 5.04 & 4.24
& 1.72 & 0.97
& 1.49 & 0.89 \\

NExT-GQA
& 10,402 & 515
& 8.94 & 4.37
& 2.12 & 0.70
& 1.63 & 0.60 \\

ActivityNet
& 8,039 & 376
& 6.65 & 4.30
& 2.41 & 0.73
& 1.62 & 0.65 \\

\midrule

MMVU
& 8,689 & 474
& 7.70 & 4.33
& 2.16 & 1.78
& 1.93 & 1.67 \\

VideoMME
& 69,750 & 3,060
& 18.72 & 4.80
& 10.03 & 1.41
& 8.17 & 1.12 \\

LongVideoBench
& 92,160 & 13,950
& 20.65 & 7.89
& 16.59 & 4.31
& 11.57 & 3.04 \\

MVBench
& 1,449 & 268
& 4.66 & 4.24
& 0.59 & 1.31
& 0.49 & 1.22 \\

MVP
& 14,625 & 1,440
& 7.77 & 4.41
& 2.96 & 2.34
& 1.88 & 1.79 \\

VideoMMMU
& 89,160 & 19,137
& 21.54 & 12.99
& 20.62 & 8.02
& 12.03 & 3.74 \\

\midrule

\textbf{Macro average}
& \textbf{33,113} & \textbf{4,386}
& \textbf{11.30} & \textbf{5.73}
& \textbf{6.58} & \textbf{2.40}
& \textbf{4.53} & \textbf{1.63} \\

\bottomrule
\end{tabular}
}
\vspace{-2mm}
\end{table*}

Averaged across all nine benchmarks, reducing the frame rate from 2.0 to
0.1 fps decreases the number of visual tokens by 86.8\%.
End-to-end latency decreases by 63.6\%, corresponding to a
2.74$\times$ speedup, while GPU inference latency decreases by 63.9\%.
Peak GPU memory usage is reduced by 49.3\%.
These results show that the reduction in sampled visual observations translates
into substantial practical inference savings rather than only a smaller nominal
frame count.
MVBench is the only benchmark for which latency increases at 0.1 fps.
This behavior is primarily associated with longer output generation rather
than increased visual processing.
At 0.1\,fps, the model generates 43.3 output tokens on average, compared with only 8.8 output tokens on average at 2.0\,fps. Despite the shorter visual input, the longer generation increases end-to-end
latency from 0.59 to 1.31 seconds and GPU inference latency from 0.49 to
1.22 seconds.

\paragraph{Measurement details.}
Visual-token counts are computed from the processor-provided video and image
grids after spatial merging.
End-to-end latency includes video loading, frame sampling, preprocessing,
host-to-device transfer, and generation.
GPU inference latency measures synchronized \texttt{model.generate()} time.
Peak GPU memory is measured using
\texttt{torch.cuda.max\_memory\_allocated()}.

\section{Comparison with Inference-Time Selection Methods}
\label{app:selection_methods}

We compare SAVER with two representative inference-time efficiency
methods: FastVID~\citep{shen2026fastvid} and
WFS-SB~\citep{Chen_2026_CVPR}.
Both methods reduce inference cost by pruning visual tokens or
selecting informative frames from richer visual inputs.
In contrast, SAVER improves sparse-frame reasoning through
dense-to-sparse post-training and operates directly under a
prescribed sparse-input budget without dense pre-scanning
or additional selection at inference time.

\begin{table*}[t]
\centering
\caption{
\textbf{Comparison with inference-time frame/token selection methods.}
We evaluate Qwen3.5-2B, SAVER, FastVID, and WFS-SB on one temporal-grounding
benchmark and two video-QA benchmarks.
SAVER and the selection-based approaches target complementary aspects of
efficient video reasoning, and combining SAVER with WFS-SB yields the strongest
overall performance.
The average is computed over the three reported benchmarks.
}
\label{tab:selection_comparison}
\vspace{-2mm}

\begin{tabular}{lcccc}
\toprule
\textbf{Method}
& \makecell{\textbf{ActivityNet}\\\textbf{TG}}
& \makecell{\textbf{LongVideoBench}\\\textbf{VQA}}
& \makecell{\textbf{MMVU}\\\textbf{VQA}}
& \textbf{Avg.} \\
\midrule

Qwen3.5-2B (0.1 fps)
& 20.7
& 51.3
& 57.0
& 43.0 \\

Qwen3.5-2B + FastVID
& 31.1
& 51.0
& 59.3
& 47.1 \\

Qwen3.5-2B + WFS-SB
& 30.0
& 56.6
& 60.3
& 49.0 \\

SAVER (0.1 fps)
& 37.2
& 56.0
& 59.7
& 51.0 \\

\rowcolor{blue!4}
SAVER + WFS-SB
& \textbf{42.1}
& \textbf{62.3}
& \textbf{60.7}
& \textbf{55.0} \\

\bottomrule
\end{tabular}

\vspace{-2mm}
\end{table*}
Table~\ref{tab:selection_comparison} shows that adaptive selection and SAVER
improve efficiency from different directions.
FastVID and WFS-SB improve the quality of the visual input by selecting or
retaining informative observations from a richer candidate pool, whereas
SAVER improves the model itself so that it can reason more reliably from a
fixed sparse input.
Under the same Qwen3.5-2B backbone, SAVER at 0.1 fps achieves an average score
of 51.0, compared with 47.1 for FastVID and 49.0 for WFS-SB across the three
reported benchmarks.

The two approaches are also naturally compositional.
When a richer candidate pool is available, WFS-SB can first select informative
frames, after which SAVER processes the resulting sparse observations.
Combining SAVER with WFS-SB improves the average performance from 51.0 to
55.0, including gains from 37.2 to 42.1 on ActivityNet and from 56.0 to 62.3
on LongVideoBench.
This result indicates that training-time sparse robustness and inference-time
adaptive selection are complementary rather than competing approaches.

\paragraph{Practical inference cost.}

We further compare the practical inference cost of \textsc{SAVER}, FastVID, and WFS-SB under matched profiling settings, using the Qwen3.5-2B backbone and averaging over the three benchmarks in Table~\ref{tab:selection_comparison}. Table~\ref{tab:selection_efficiency} reports visual-token count, peak GPU
memory, GPU inference latency and end-to-end latency.
\begin{table*}[t]
\centering
\caption{
\textbf{Practical inference cost of SAVER and inference-time selection methods.}
Lower values indicate lower inference cost.
SAVER directly operates on sparse inputs, while FastVID and WFS-SB introduce
additional inference-time processing for pruning or frame selection.
}
\label{tab:selection_efficiency}
\vspace{-2mm}

\begin{tabular}{lccccc}
\toprule
\textbf{Method}
& \makecell{\textbf{Visual}\\\textbf{tokens} $\downarrow$}
& \makecell{\textbf{Peak memory}\\\textbf{(GiB)} $\downarrow$}
& \makecell{\textbf{GPU inference}\\\textbf{(s)} $\downarrow$}
& \makecell{\textbf{E2E latency}\\\textbf{(s)} $\downarrow$} \\
\midrule

\rowcolor{blue!4}
SAVER
& \textbf{4,933}
& \textbf{5.51}
& \textbf{1.79} 
& \textbf{2.27} \\

FastVID
& 5,292
& 13.39
& 2.22
& 4.42 \\

WFS-SB
& 5,393
& 6.96
& 1.94
& 6.36 \\

\bottomrule
\end{tabular}

\vspace{-2mm}
\end{table*}
The practical-cost comparison further highlights the different inference
trade-offs.
SAVER processes fewer visual tokens and uses less peak GPU memory than the two
selection-based methods, while also requiring lower GPU inference latency and
end-to-end latency under the matched profiling
setting.
This advantage follows from the different inference pipeline: SAVER improves
sparse-input capability during post-training and therefore does not require
dense pre-scanning, intermediate feature extraction, or an additional
selection stage at inference time.

FastVID and WFS-SB instead spend additional inference-time computation to
identify informative visual content from a richer candidate representation.
Such processing can be worthwhile when a richer input pool is available, as
demonstrated by the improved performance of SAVER + WFS-SB in
Table~\ref{tab:selection_comparison}.
Taken together, these results suggest that SAVER provides a complementary
training-time approach to efficient video reasoning: it reduces the model's
dependence on dense observations, while adaptive selection methods can further
improve the sparse input when additional inference-time computation is
available.

\section{Effect of Reference Reward Weight}
\label{app:ref_reward_weight}

We study the sensitivity of SAVER to the reference reward coefficient
$\lambda_{\mathrm{ref}}$.
This analysis examines whether the observed gains can be explained simply by
increasing the relative contribution of the reference reward during
optimization.
All results use Qwen3.5-2B and are evaluated at 0.1 fps.

\begin{table}[t]
\centering
\caption{
\textbf{Effect of the reference reward weight
$\lambda_{\mathrm{ref}}$.}
Temporal-grounding results are averaged over three benchmarks and video-QA
results over six benchmarks.
}
\label{tab:ref_reward_weight}
\vspace{-2mm}
\begin{tabular}{c|cc}
\toprule
$\lambda_{\mathrm{ref}}$
& \makecell{\textbf{TG}\\\textbf{mIoU}}
& \makecell{\textbf{QA}\\\textbf{Avg.}} \\
\midrule
0.0 & 27.4 & 48.7 \\
0.5 & 29.2 & 50.0 \\
1.0 & 30.0 & \textbf{50.8} \\
2.0 & \textbf{30.1} & 50.2 \\
\bottomrule
\end{tabular}
\vspace{-2mm}
\end{table}

As shown in Table~\ref{tab:ref_reward_weight}, reward weighting affects
performance, but the gains are not explained by monotonically increasing the
reference reward weight.
Starting from $\lambda_{\mathrm{ref}}=0$, introducing the reference objective
substantially improves both temporal grounding and video-QA.
Performance improves as $\lambda_{\mathrm{ref}}$ increases to 1.0, whereas
further increasing it to 2.0 yields only a marginal improvement in temporal
grounding (30.0 to 30.1 mIoU) while reducing video-QA accuracy from 50.8 to
50.2. These results indicate that the reference reward contributes a distinct
cross-view alignment objective, while $\lambda_{\mathrm{ref}}$ controls the
strength of this objective during optimization.
The benefit therefore comes from both the dense-to-sparse alignment signal and
an appropriate balance among reward components, rather than from simply
increasing the magnitude of a single reward term.

\section{Effect of Grounding Post-Training Data Scale}

We study the effect of the grounding post-training data scale in Table~\ref{tab:grounding_data_scale}.
Time-R1 contains approximately 2500 temporal grounding training examples, while our main experiments use only a randomly sampled subset of 1250 examples.
We observe that increasing the amount of grounding supervision initially improves both temporal grounding and video-QA performance.
However, the gains saturate beyond 1250 examples, and using the full 2500-example training set slightly degrades performance under sparse inference.
We hypothesize that excessive grounding-focused post-training may over-specialize the model toward narrow localization patterns and reduce the diversity of temporal reasoning behaviors.
Overall, these results suggest that effective dense-to-sparse reasoning transfer can emerge from a relatively small amount of grounding supervision.

\begin{table}[t]
\centering
\small
\setlength{\tabcolsep}{6pt}
\caption{
\textbf{Effect of grounding post-training data scale.}
We vary the number of Time-R1 grounding training examples used during post-training. Results are evaluated under 0.1-fps inference.
}
\label{tab:grounding_data_scale}
\begin{tabular}{lcccc}
\toprule
\textbf{Model}
& \textbf{Size}
& \textbf{Training Samples}
& \textbf{TG Avg. mIoU}
& \textbf{Video-QA Avg. Acc.} \\
\midrule
\textsc{SAVER}
& 2B
& 0
& 18.4
& 45.9 \\

\textsc{SAVER}
& 2B
& 625
& 25.7
& 47.6 \\

\rowcolor{blue!5}
\textsc{SAVER}
& 2B
& 1250
& \textbf{30.0}
& \textbf{50.8} \\

\textsc{SAVER}
& 2B
& 1875
& 29.5
& 49.7\\

\textsc{SAVER}
& 2B
& 2500
& 28.6
& 48.2 \\
\bottomrule
\end{tabular}
\end{table}

\section{Confidence-Based Sparse-to-Dense Fallback}
\label{app:fallback}

\textsc{SAVER} is designed for direct sparse-frame inference and does not require
dense video access at test time. Nevertheless, when additional computation is
available, the sparse prediction can optionally be used to decide whether a
denser observation should be requested. We therefore study a simple
confidence-based sparse-to-dense fallback strategy.

We first perform inference using sparse frames sampled at 0.1 fps. If the
confidence of the sparse prediction falls below a threshold $\gamma$, we
rerun the same example using frames sampled at 2.0 fps and use the dense-view
prediction instead. Table~\ref{tab:fallback} reports the resulting
performance--computation trade-off for temporal grounding and video-QA.

\begin{table}[t]
\centering
\caption{
\textbf{Confidence-based sparse-to-dense fallback.}
Inference starts from the 0.1-fps sparse view, and dense 2.0-fps inference is
triggered when the sparse prediction confidence is below $\gamma$.
Increasing $\gamma$ routes more examples to dense inference and therefore
increases the average number of processed frames.
}
\label{tab:fallback}
\resizebox{\linewidth}{!}{
\begin{tabular}{l c cc cc}
\toprule
\multirow{2}{*}{\textbf{Policy}}
& \multirow{2}{*}{$\boldsymbol{\gamma}$}
& \multicolumn{2}{c}{\textbf{Temporal Grounding}}
& \multicolumn{2}{c}{\textbf{Video-QA}} \\
\cmidrule(lr){3-4}
\cmidrule(lr){5-6}
&
& \makecell{Avg.\\Frames}
& \makecell{Avg.\\mIoU}
& \makecell{Avg.\\Frames}
& \makecell{Avg.\\Acc.} \\
\midrule
Qwen3.5 (dense)
& -- & 110 & 34.9 & 139 & 51.0 \\

\midrule
\textsc{SAVER} (sparse only)
& -- & 7 & 30.0 & 38 & 50.8 \\

Fallback
& 0.3 & 8  & 29.1 & 39 & 50.5 \\

Fallback
& 0.5 & 11 & 31.0 & 43 & 50.9 \\

Fallback
& 0.7 & 47 & 38.3 & 54 & 52.0 \\

\midrule
\textsc{SAVER} (dense)
& -- & 110 & \textbf{45.4} & 139 & \textbf{54.6} \\
\bottomrule
\end{tabular}
}
\end{table}

As $\gamma$ increases, more examples are routed to dense inference, increasing
the average frame budget and generally moving performance toward the dense-view
setting. However, the fallback strategy does not improve performance
monotonically. At lower thresholds, the confidence signal is not sufficiently
calibrated to identify examples that truly require additional temporal
evidence. As a result, some correct sparse predictions are replaced by less
helpful dense predictions, leading to degraded temporal-grounding performance.

At $\gamma=0.7$, the fallback strategy improves temporal grounding from
30.0 to 38.3 average mIoU and video-QA from 50.8 to 52.0 average accuracy,
while processing substantially fewer frames than dense-only inference
(47 vs.\ 110 frames for temporal grounding and 54 vs.\ 139 frames for
video-QA). These results indicate that selective access to denser observations
can provide additional gains when sparse evidence is insufficient, but its
effectiveness depends critically on reliable confidence calibration.
We therefore treat confidence-based fallback as an optional extension rather
than part of the default \textsc{SAVER} inference protocol.

\section{Additional Qualitative Cases}
\label{app:failure_cases}

\begin{figure}[t]
    \centering
    \includegraphics[width=1\linewidth]{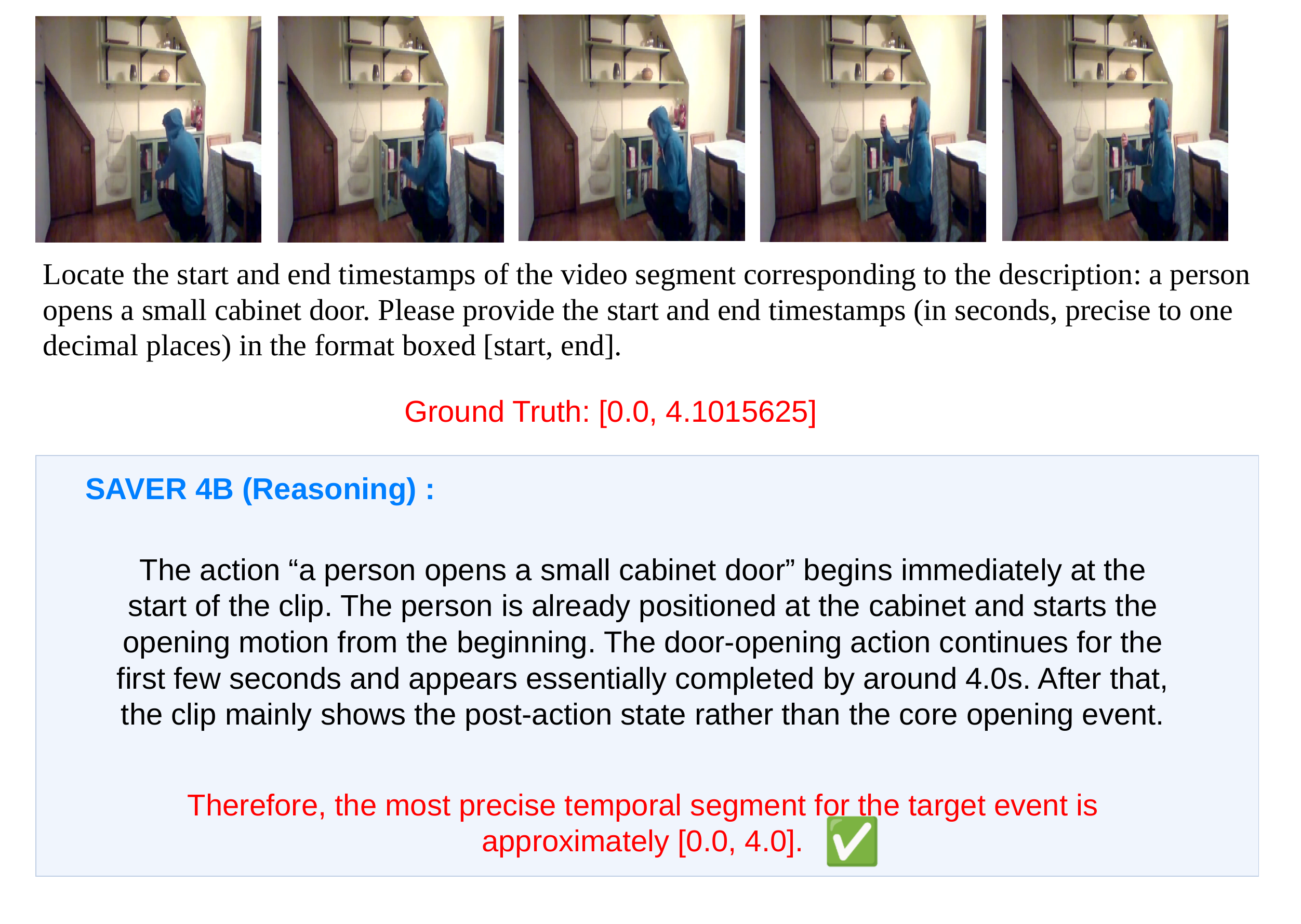}
    \caption{
    \textbf{Temporal grounding success case on Charades-STA.}
    Given the query ``a person opens a small cabinet door,'' \textsc{SAVER} correctly identifies that the action starts at the beginning of the clip and ends around 4 seconds.
    The predicted interval closely matches the ground-truth segment, showing that sparse-frame reasoning can succeed when the sampled frames preserve the key temporal evidence.
    }
    \label{fig:charades_success_case}
\end{figure}

\begin{figure}[t]
    \centering
    \includegraphics[width=1\linewidth]{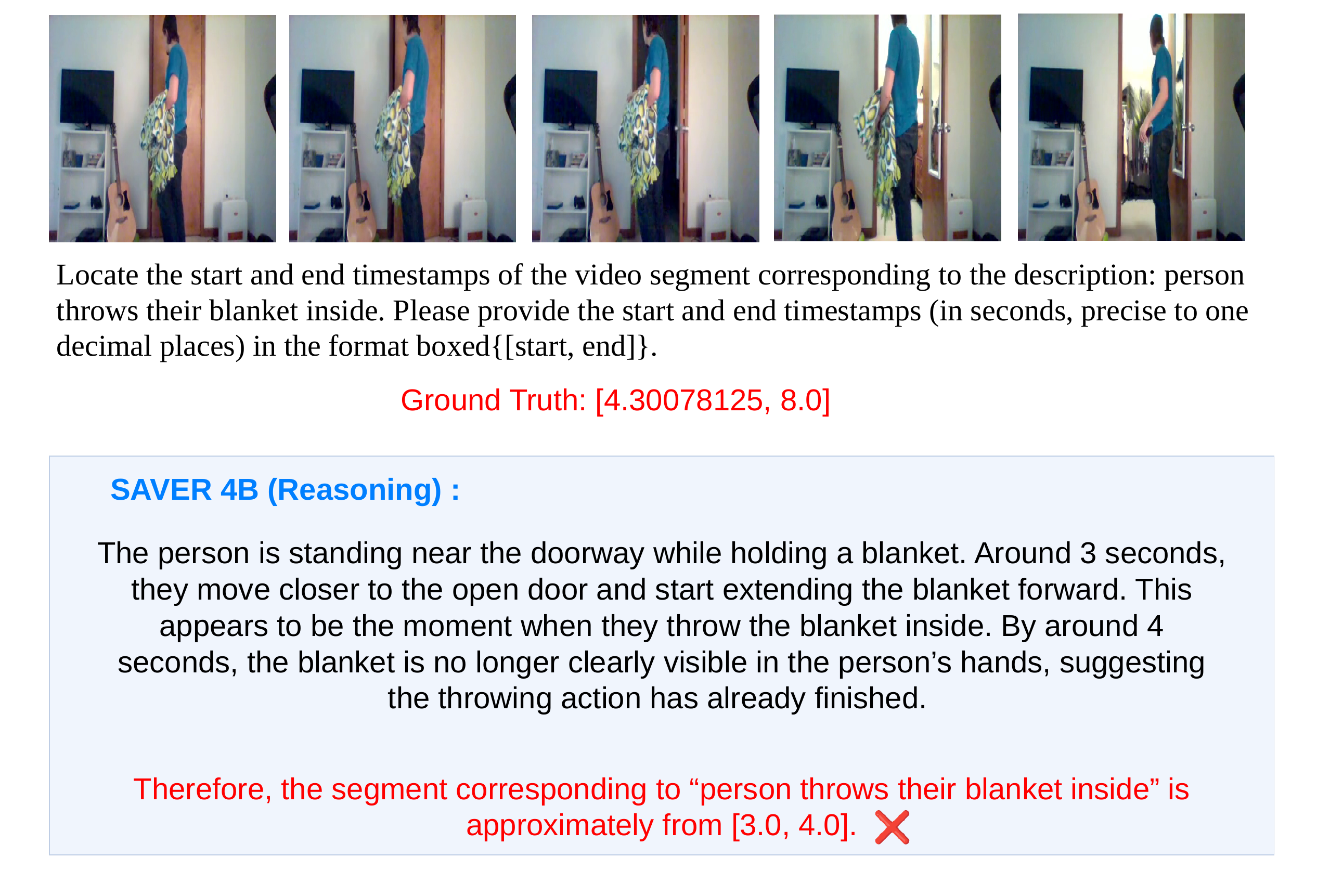}
    \caption{
    \textbf{Temporal grounding failure case on Charades-STA.}
    Given the query ``person throws their blanket inside,'' \textsc{SAVER} predicts an earlier interval than the ground truth.
    Although the reasoning identifies a plausible moment when the person moves near the doorway, the sparse observations do not provide enough evidence to precisely distinguish the target throwing action.
    This example highlights a common failure mode: sparse-frame inference can still localize an incorrect segment when the key action is ambiguous or insufficiently captured.
    }
    \label{fig:charades_failure_case}
\end{figure}

\begin{figure}[t]
    \centering
    \includegraphics[width=1\linewidth]{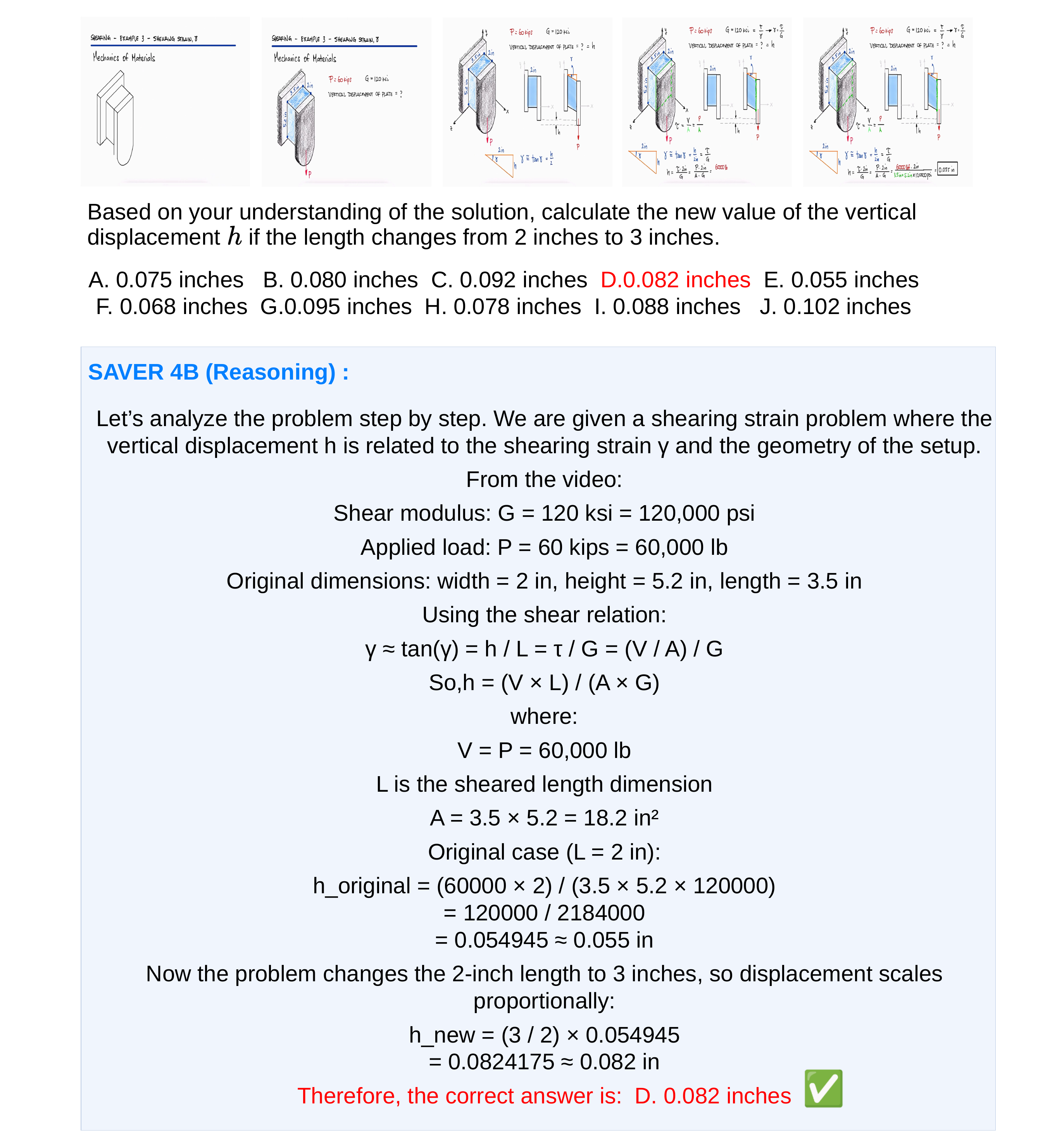}
    \caption{
    \textbf{Video-QA success case on VideoMMMU.}
    Given a mechanics question about how the vertical displacement changes when the length increases from 2 inches to 3 inches, \textsc{SAVER} correctly follows the visual solution and derives the proportional change.
    The model identifies the relevant formula, applies the new length, and selects the correct answer, showing that sparse video reasoning can succeed when the sampled frames retain the necessary visual and textual evidence.
    }
    \label{fig:videommmu_success_case}
\end{figure}

\begin{figure}[t]
    \centering
    \includegraphics[width=1\linewidth]{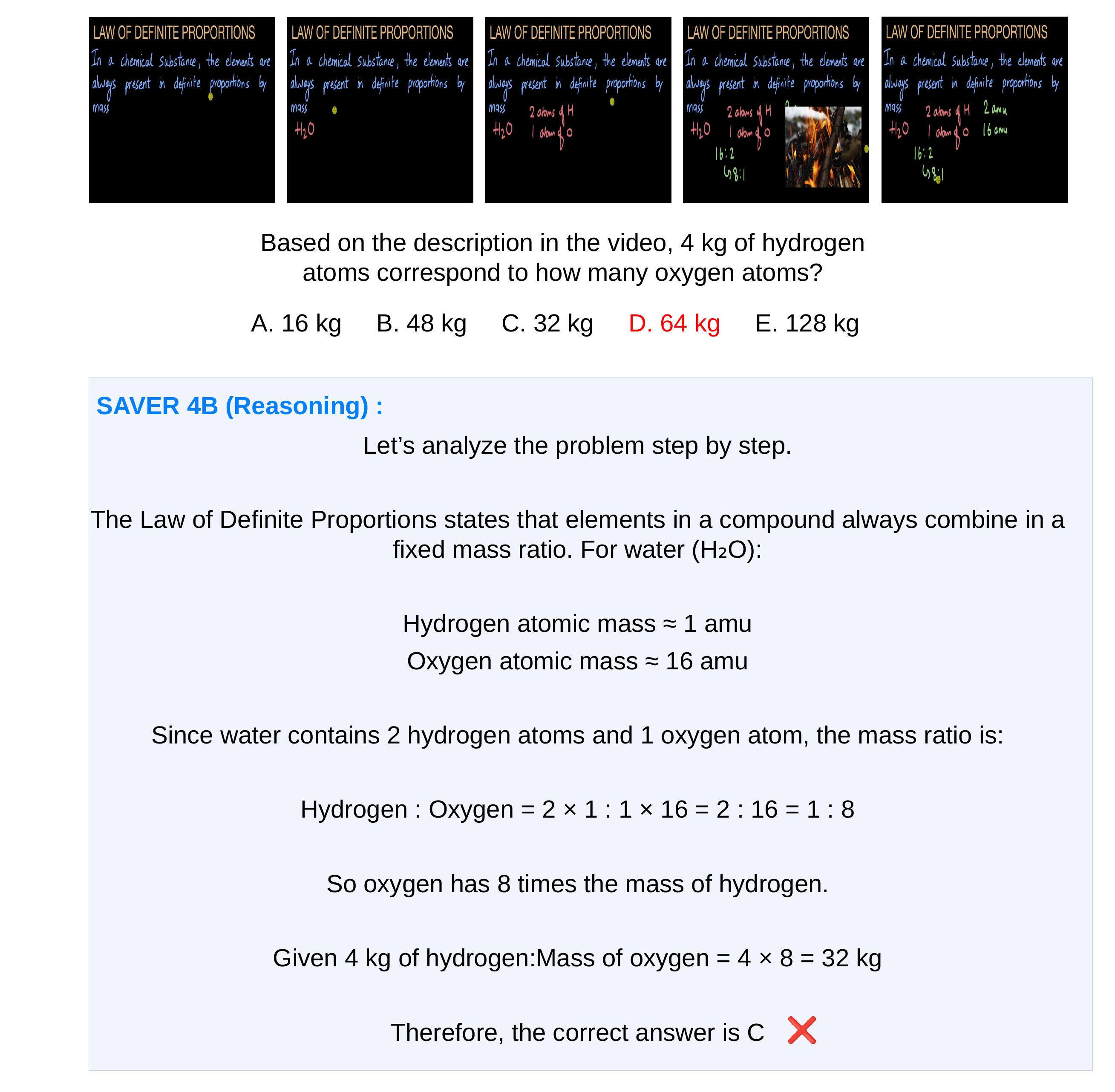}
\caption{
    \textbf{Video-QA failure case on VideoMMMU.}
    Given a chemistry question about the mass of oxygen corresponding to 4\,kg of hydrogen atoms, \textsc{SAVER} applies the standard hydrogen-to-oxygen mass ratio in water (1:8) and selects C (32\,kg), whereas the ground-truth answer is D (64\,kg), which corresponds to a 1:16 mass relation.
    Although the arithmetic is internally consistent, the model relies on a default prior assumption instead of the relation presented in the video.
    This example shows that sparse video reasoning can still fail when the reasoning chain is built on an incorrect premise that is not grounded in the visual evidence.
    }
    \label{fig:videommmu_failure_case}
\end{figure}

We provide additional qualitative examples for both temporal grounding and video question answering under the extremely sparse inference setting of 0.1 fps. Figures~\ref{fig:charades_success_case} and~\ref{fig:charades_failure_case} show Charades-STA grounding cases, where \textsc{SAVER} predicts temporal segments from sparse frames.
Figures~\ref{fig:videommmu_success_case} and~\ref{fig:videommmu_failure_case} show VideoMMMU cases, where \textsc{SAVER} must solve reasoning-intensive video questions from sparse observations.
Together, these examples show that \textsc{SAVER} can preserve task-relevant evidence when the sampled frames contain sufficient cues, but may still fail when the sparse observations miss critical information or when the reasoning chain contains an incorrect intermediate assumption.

\end{document}